\documentclass[10pt,twocolumn,letterpaper]{article}

\usepackage{cvpr}      % To produce the REVIEW version
\usepackage{tikz}
\usetikzlibrary{positioning, shapes, arrows.meta, calc, backgrounds, fit, shadows.blur}
\usepackage{xcolor}% --- inline annotations
\makeatletter\renewcommand\paragraph{\@startsection{paragraph}{4}{\z@}
  {.5em \@plus1ex \@minus.2ex}{-.5em}{\normalfont\normalsize\bfseries}}\makeatother

\definecolor{cvprblue}{rgb}{0.21,0.49,0.74}
\usepackage[pagebackref,breaklinks,colorlinks,allcolors=cvprblue]{hyperref}

\usepackage{tabularx, multirow, stfloats, bbding, makecell, soul, colortbl}
\newcolumntype{C}{>{\centering \arraybackslash}X}

\usepackage{algorithm}
\usepackage{algorithmic}

\def\paperID{} % *** Enter the Paper ID here
\def\confName{CVPR}
\def\confYear{2026}

\title{V-GRPO: Online Reinforcement Learning for Denoising Generative Models Is Easier than You Think}

\author{
\textbf{Bingda Tang$^{1,2}$\thanks{Equal contribution. Code: \url{https://github.com/tang-bd/v-grpo}.} \quad Yuhui Zhang$^{1*}$ \quad Xiaohan Wang$^{1}$ \quad Jiayuan Mao$^{3,4}$} \\ 
\textbf{Ludwig Schmidt$^{1}$ \quad Serena Yeung-Levy$^{1}$} \\
$^{1}$Stanford University \quad $^{2}$Tsinghua University \quad $^{3}$Amazon FAR \quad $^{4}$University of Pennsylvania
}

\begin{document}
\maketitle
\begin{abstract}

Aligning denoising generative models with human preferences or verifiable rewards remains a key challenge. While policy-gradient online reinforcement learning (RL) offers a principled post-training framework, its direct application is hindered by the intractable likelihoods of these models. 
Prior work therefore either optimizes an induced Markov decision process (MDP) over sampling trajectories, which is stable but inefficient, or uses likelihood surrogates based on the diffusion evidence lower bound (ELBO), which have so far underperformed on visual generation.  Our key insight is that the ELBO-based approach can, in fact, be made both stable and efficient. By reducing surrogate variance and controlling gradient steps, we show that this approach can beat MDP-based methods. 
To this end, we introduce Variational GRPO (V-GRPO), a method that integrates ELBO-based surrogates with the Group Relative Policy Optimization (GRPO) algorithm, alongside a set of simple yet essential techniques. Our method is easy to implement, aligns with pretraining objectives, and avoids the limitations of MDP-based methods. \textbf{V-GRPO achieves state-of-the-art performance in text-to-image synthesis, while delivering a $\mathbf{2}\boldsymbol{\times}$ speedup over MixGRPO and a $\mathbf{3}\boldsymbol{\times}$ speedup over DiffusionNFT.}

\end{abstract}    
\section{Introduction}
\label{sec:intro}

The recent success of online reinforcement learning (RL) for post-training large language models (LLMs)~\cite{deepseekmath, deepseekr1} has rekindled interest in applying similar techniques to denoising generative models~\cite{diffusion, ddpm, flowmatching}. Such post-training is crucial for aligning pretrained models with human preferences~\cite{pickscore, hpsv2, imagereward, unifiedreward} or verifiable rewards~\cite{geneval, paddleocr}.

While policy gradient methods such as Proximal Policy Optimization (PPO)~\cite{ppo} and Group Relative Policy Optimization (GRPO)~\cite{deepseekmath} have become prominent in the LLM literature, they are difficult to apply to denoising generative models because they require access to exact likelihoods, which are generally intractable for these models.

\begin{table}[t!]
    \centering
    \renewcommand{\arraystretch}{1.1}
    \vspace{-3mm}
    \footnotesize
    \caption{\textbf{V-GRPO delivers state-of-the-art performance with substantial speedup over baselines.} Results are averaged across all evaluated reward functions and reported on the test sets. Full results are presented in~\cref{tab:main-flux} and~\cref{tab:main-sd}, and the experimental setup is described in~\cref{sec:setup} and~\cref{sec:impl}.}
    \vspace{-0.5em}
        \begin{tabular}{lccccc}
            \toprule
            Method & \#Steps &
$\text{NFE}_{\pi^{\boldsymbol{\theta}_{\text{old}}}}$& $\text{NFE}_{\pi^{\boldsymbol{\theta}}}$ & Reward \\
            \midrule
            FLUX.1-dev & — & —  & — & 1.25 \\
            + BranchGRPO & 300 & 13.68 & 13.68 & 1.40 \\
            + MixGRPO & 300 & 25 & 4 & 1.41\\
            + V-GRPO & 150 & 16 + 4 & 4 & 1.42 \\
            + V-GRPO & 300 & 16 + 4 & 4 & \textbf{1.45} \\
            \midrule
            SD 3.5 M (w/o CFG) & — & —  & — & 0.95 \\
            + DiffusionNFT & 1.7K & 40 + 40 & 40& \textbf{1.71} \\
            + V-GRPO & 580 & 40 + 6.9 & 6.9 & \textbf{1.71} \\
    
            \bottomrule
            \end{tabular}
\vspace{-2em}
\label{tab:teaser}
\end{table}

A classic workaround is to impose a stochastic sampling process and frame generation as a Markov decision process (MDP)~\cite{dpok, ddpo, dancegrpo, flowgrpo, mixgrpo, branchgrpo}. From this perspective, the joint probability of a sampling trajectory factorizes into a sequence of reverse transition kernels, each given by a tractable Gaussian. This decomposition induces a sequential state-action space, enabling the direct application of policy gradient methods. The theoretical validity of this approach rests on the fact that the likelihood of the final output is recovered by marginalizing the joint distribution over the full trajectory, which allows the training objective to be estimated via Monte Carlo integration over rollouts~\cite{shortcut}.

While conceptually sound, this formulation presents several limitations. First, the MDP objective suffers from slow convergence, resulting in inefficient training. Second, modeling generation as an MDP confines sampling to first-order stochastic differential equation (SDE) discretizations, precluding the use of more efficient higher-order or ordinary differential equation (ODE) solvers. Finally, binding optimization to rollout transition kernels creates a tight coupling between the two stages, limiting implementation flexibility.

These limitations invite increasingly elaborate designs to patch the inefficiency and inflexibility of the MDP framework. For instance, MixGRPO~\cite{mixgrpo} introduces a hybrid ODE–SDE sampling scheme with a sliding-window schedule, while BranchGRPO~\cite{branchgrpo} restructures sampling into a branching tree. Both yield notable improvements, but at the cost of substantially higher algorithmic complexity and more hyperparameters.

A simpler yet often overlooked approach is to revisit the variational roots of diffusion models: adopting pretraining objectives closely connected to the diffusion evidence lower bound (ELBO) as tractable surrogates for the model log-likelihood within policy gradient methods. Despite its promise on standard RL benchmarks, this approach has been reported to significantly underperform in visual generation tasks \cite{ddpo, fpo}. In this work, we demonstrate that this gap is not fundamental. 

To close this gap, we present Variational GRPO (V-GRPO), a method that integrates ELBO-based surrogates with the GRPO algorithm. Our key insight is that a carefully chosen set of surrogate variance reduction and gradient step regularization techniques, while simple individually, prove essential for stable training and superior performance. Together, these yield a method that is easy to implement, aligns with pretraining, and avoids the limitations of MDP-based approaches. 

The results are compelling: for multi-reward text-to-image synthesis, \textbf{V-GRPO achieves state-of-the-art performance and runs $\mathbf{2\boldsymbol{\times}}$ faster than the leading MDP-based baseline, MixGRPO}. In a multi-stage and multi-reward setting, \textbf{V-GRPO matches the performance of DiffusionNFT~\cite{diffusionnft} while delivering a $\mathbf{3\boldsymbol{\times}}$ speedup}.

Our findings prove that properly stabilized ELBO-based methods are not only superior to complex MDP approaches but also competitive across other methods. They offer a new default for post-training denoising generative models. We hope that this work will provide useful insights and guidance, encouraging future research along this direction.
\section{Related Work}
\label{sec:related}

Extensive research has explored applying RL to denoising generative models such as diffusion~\cite{diffusion, ddpm} and flow matching models~\cite{flowmatching}, with existing approaches broadly divided into offline and online paradigms.

Offline methods commonly employ ELBO-based surrogates to approximate model log-likelihoods, as in reward-weighted regression (RWR)~\cite{rwr} and Direct Preference Optimization (DPO)~\cite{dpo, diffusiondpo}. Despite their simplicity, offline RL approaches are limited by distributional shift and the restricted coverage of static datasets.

In contrast, online approaches typically formulate the stochastic sampling process as an MDP, optimizing reverse transition kernels over the induced state-action space via policy gradient methods~\cite{shortcut, ddpo, dpok}. Recent work has extended this paradigm through GRPO-based variants and flow matching models~\cite{dancegrpo, flowgrpo}, alongside more sophisticated algorithmic designs addressing both theoretical and practical limitations~\cite{cps, mixgrpo, branchgrpo, grpoguard, tempflowgrpo}.

While some online methods employ ELBO-based surrogates, DDPO~\cite{ddpo} and FPO~\cite{fpo} have shown these to underperform on visual generation tasks. In this work, we revisit this simple approach and demonstrate that this limitation is not fundamental: a set of simple yet effective techniques unlocks its full potential, achieving state-of-the-art performance with significantly improved training efficiency. Concurrent with our work, Advantage Weighted Matching (AWM)~\cite{awm} also explores ELBO-based surrogates and demonstrates their underexplored potential, yet our work offers a more comprehensive study with stronger empirical validation.

To circumvent likelihood approximation altogether, DiffusionNFT~\cite{diffusionnft} foregoes standard policy gradient framework in favor of contrasting positive and negative policies, achieving impressive results. Similar to our method, it builds upon the pretraining objectives. However, it requires maintaining two sets of model weights, which introduces additional overhead.
\section{Preliminaries}
\label{sec:preliminaries}

\subsection{Denoising Generative Models}

In denoising generative models, such as diffusion~\cite{diffusion, ddpm} and flow matching models~\cite{flowmatching}, a forward process $\mathbf{z}_t$ is defined for $t\in [0, 1]$. This process gradually transforms data samples $\mathbf{x} \sim \pi_0$ from a target distribution into noise samples $\boldsymbol{\epsilon} \sim \pi_1$, where $\pi_1$ is a tractable prior (typically $\mathcal{N}(\mathbf{0},\mathbf{I})$). A convenient and widely adopted parameterization for this process is the linear interpolation
\begin{equation}
    \mathbf{z}_t = a_t\mathbf{x} + b_t\boldsymbol{\epsilon},
\end{equation}
where $a_t, b_t$ are method-specific schedule coefficients.

The goal is to learn a parameterized neural network $\texttt{NN}^{\boldsymbol{\theta}}_t$ that approximates the reverse dynamics by minimizing a suitably weighted regression objective
\begin{equation}
\mathcal{L}_w(\boldsymbol{\theta})
= \mathbb{E}_{t, \pi_0(\mathbf{x}), \pi_1(\boldsymbol{\epsilon})}
\Big[w_t\Big\| \texttt{NN}^{\boldsymbol{\theta}}_t(\mathbf{z}_t) - \mathbf{r}_t(\mathbf{x}, \boldsymbol{\epsilon})  \Big\|_2^2\Big],
\label{eq:loss}
\end{equation}
where $w_t$ is a weighting function and $\mathbf{r}_t$ denotes the regression target, often defined by another linear interpolation
\begin{equation}
    \mathbf{r}_t = c_t\mathbf{x} + d_t\boldsymbol{\epsilon},
\end{equation}
where $c_t, d_t$ are also method-specific coefficients. Popular choices of $\mathbf{r}_t$ include $\mathbf{x}$-prediction~\cite{salimans2022progressive, jit}, $\boldsymbol{\epsilon}$-prediction~\cite{ddpm, score, vdm, maximum}, and $\mathbf{v}$-prediction ($\mathbf{v} = \boldsymbol{\epsilon} - \mathbf{x}$)~\cite{flowmatching, rf, salimans2022progressive}.

Given a trained \(\texttt{NN}^{\boldsymbol{\theta}}_t\) and an initial noise sample \(\mathbf{x}\), first-order sampling proceeds by iterative denoising. In a discretized time schedule \(\{t_i\}\) this can be written as
\begin{equation}
    \mathbf{x}_{t_{i-1}} = \alpha_{t_i}\mathbf{x}_{t_i} + \beta_{t_i}\,\texttt{NN}^{\boldsymbol{\theta}}_{t_i}(\mathbf{x}_{t_i})
    + \sigma_{t_i}\,\boldsymbol{\tilde{\epsilon}}_{t_i},
    \label{eq:reverse}
\end{equation}
where \(\alpha_{t_i},\beta_{t_i},\sigma_{t_i}\) are sampler-specific schedule parameters. The noise term \(\boldsymbol{\tilde{\epsilon}}_{t_i}\sim\mathcal{N}(\mathbf{0},\mathbf{I})\) introduces stochasticity in SDE samplers, while ODE samplers correspond to the deterministic case with $\sigma_{t_i} \equiv 0$. High-order samplers can be formulated analogously with extra schedules.

\subsection{Group Relative Policy Optimization}

Compared to PPO~\cite{ppo}, GRPO~\cite{deepseekmath} removes the value function and estimates the advantage baseline using a group-relative approach. For an input $c$, the behavior policy $\pi^{\boldsymbol{\theta}_\text{old}}$ generates a group of outputs $\{\mathbf{o}_i\}_{i=1}^G$, and the advantage of the $i$-th output is obtained by normalizing its reward $R_i$ with the group-level mean and standard deviation of $\{R_i\}_{i=1}^G$:
\begin{equation}
    A_i = \frac{R_i - \text{mean}(\{R_i\}_{i=1}^G)}{\text{std}(\{R_i\}_{i=1}^G)}.
\end{equation}

The model is then updated by maximizing the clipped surrogate objective
\begin{equation}
\begin{aligned}
\mathcal{J}^\text{GRPO}(\boldsymbol{\theta}) = &\,\, \mathbb{E}_{\{\mathbf{o}_i\}_{i=1}^G \sim \pi^{\boldsymbol{\theta}_{\text{old}}}} 
\bigg[ \frac{1}{G} \sum_{i=1}^G \min\big( \mathbf{\rho}^{\boldsymbol{\theta}}_{i} A_i, \\
& \,\, \text{clip}( \mathbf{\rho}^{\boldsymbol{\theta}}_{i}, 1-\epsilon, 1+\epsilon) A_i \big) \bigg].
\end{aligned}
\label{eq:grpo}
\end{equation}
Here, the importance ratio is defined as $\rho^{\boldsymbol{\theta}}_i = \frac{\pi^{\boldsymbol{\theta}}(\mathbf{o}_i | c)}{\pi^{\boldsymbol{\theta}_{\text{old}}}(\mathbf{o}_i | c)}$, while $\epsilon$ denotes the clipping range for this ratio. 

Our method treats multi-step generation as an atomic action within the RL framework, directly optimizing the policy based on the marginal likelihood surrogate of the final output $\pi^{\boldsymbol{\theta}}(\mathbf{o}_i|c)$. In contrast, MDP-based approaches~\cite{dpok, ddpo, dancegrpo, flowgrpo, mixgrpo, branchgrpo} and prior LLM work~\cite{deepseekr1, deepseekmath} model generation as a sequence of individual actions, optimizing per-step objectives that average the clipped surrogate loss across all transitions in the trajectory.

\begin{algorithm*}[htp]
\caption{V-GRPO}
\label{alg:vgrpo}
\begin{algorithmic}[1]
\REQUIRE Initial policy $\boldsymbol{\theta}$, reward functions ${\{R^k\}_{k=1}^K}$, prompt dataset $\mathcal{D}$, hyperparameters $\beta, \epsilon, \eta, N_{\text{MC}}, \dots$
\FOR{$\text{iteration} = 1, \dots, M$}
    \STATE Sample $N$ batches of prompts $\mathcal{D}_{\text{bN}} \subset \mathcal{D}$.
    \STATE Update the behavior policy $\boldsymbol{\theta}_{\text{old}} \leftarrow \boldsymbol{\theta}$.
    \STATE Generate $G$ outputs $\{\mathbf{o}_i\}_{i=1}^G \sim \pi^{\boldsymbol{\theta}_\text{old}}(\cdot|c)$ for each prompt $c \in \mathcal{D}_{\text{bN}}$.
    \STATE Store a set of timestep-noise pairs $\{(t_j, \boldsymbol{\epsilon}_j)\}_{j=1}^{N_{\text{MC}}}$, where timesteps are drawn via \textcolor{Plum}{stratified sampling} rather than uniform sampling. This set is \textcolor{Plum}{shared across all outputs for each prompt $c$}, rather than sampled independently per output $\mathbf{o}_i$.
    \FOR{each output $\mathbf{o}_i$ generated from each prompt $c$}
        \STATE Compute rewards from each function $\{R_{i}^k\}_{k=1}^{K}$.
        \STATE  Aggregate advantages $A_i = \text{Aggregate}(\{R_{i}^k\})$, and apply \textcolor{ForestGreen}{soft-clipping} $A^{\text{soft}}_{_i} = \eta \cdot \tanh(\frac{1}{\eta}A_i)$.
        \STATE  Compute ELBO-based surrogates with \textcolor{Plum}{adaptive loss weighting} $\hat{\mathcal{L}}^{\text{adaptive}}(\boldsymbol{\theta_{\text{old}}}\mid \mathbf{o}_i, c) = \frac{1}{N_{\text{MC}}}\sum_{j=1}^{N_{\text{MC}}} \ell^{\boldsymbol{\theta}_{\text{old}}}_i(t_j, \boldsymbol{\epsilon}_j)$.
    \ENDFOR
    \FOR{$\text{gradient step} = 1, \dots, N$}
        \STATE Sample a single batch $\mathcal{D}_{\text{b}} \subset \mathcal{D}_{\text{bN}}$.
        \FOR{each sample $\mathbf{o}_i$ in the batch $\mathcal{D}_{\text{b}}$}
            \STATE Compute $\hat{\mathcal{L}}^{\text{adaptive}}(\boldsymbol{\theta}\mid \mathbf{o}_i, c)$ with \textcolor{Plum}{adaptive loss weighting} using the current policy $\boldsymbol{\theta}$ and stored $\{(t_j, \boldsymbol{\epsilon}_j)\}_{j=1}^{N_{\text{MC}}}$.
            \STATE Calculate the importance ratio $\rho^{\boldsymbol{\theta}}_i = \exp(-\hat{\mathcal{L}}^{\text{adaptive}}(\boldsymbol{\theta}\mid \mathbf{o}_i, c) + \hat{\mathcal{L}}^{\text{adaptive}}(\boldsymbol{\theta}_{\text{old}}\mid \mathbf{o}_i, c))$.
            \STATE Compute GRPO objective with \textcolor{ForestGreen}{importance ratio clipping} $\mathcal{J}^{\text{GRPO}}_{i}(\boldsymbol{\theta}) = \min(\rho^{\boldsymbol{\theta}}_i A^{\text{soft}}_i, \text{clip}(\rho^{\boldsymbol{\theta}}_i, 1 - \epsilon, 1+ \epsilon) A^{\text{soft}}_i)$.
            \STATE Compute the \textcolor{ForestGreen}{KL penalty} $\mathbb{D}^{\text{simple}}_i(\pi^{\boldsymbol{\theta}}\Vert \pi^{\boldsymbol{\theta}_{\text{old}}})$.
        \ENDFOR
        \STATE Update the policy $\boldsymbol{\theta}\leftarrow \text{Optimizer}\big(\boldsymbol{\theta}, \nabla_{\boldsymbol{\theta}} \sum (\mathcal{J}^{\text{GRPO}}_i\big(\boldsymbol{\theta}\big) - \beta \cdot \mathbb{D}^{\text{simple}}_i(\pi^{\boldsymbol{\theta}}\Vert \pi^{\boldsymbol{\theta}_{\text{old}}}))\big)$.
    \ENDFOR
\ENDFOR
\end{algorithmic}
\end{algorithm*}
\section{Approach}
\label{sec:approach}

\subsection{Overview}
\label{sec:overview}

Unlike MDP-based methods that rely on rollout transition kernels, V-GRPO directly plugs in pretraining objectives closely connected to the diffusion ELBO as surrogates for model log-likelihoods within the GRPO algorithm. Specifically, we replace $\log \pi^{\boldsymbol{\theta}}(\mathbf{o}_i \mid c)$ with a surrogate obtained by conditioning the negative pretraining objective \cref{eq:loss} on the generated output $\mathbf{o}_i$ and prompt $c$:
\begin{equation}
\begin{aligned}
 \log \pi^{\boldsymbol{\theta}}(\mathbf{o}_i \mid c) \leftarrow &   -\mathcal{L}_w(\boldsymbol{\theta}\mid \mathbf{o}_i, c) \\
= & -\mathbb{E}_{t, \pi_1(\boldsymbol{\epsilon})}
\Big[w_t\Big\| \texttt{NN}^{\boldsymbol{\theta}}_t(\mathbf{z}_t, c) - \mathbf{r}_t(\mathbf{o}_i, \boldsymbol{\epsilon})  \Big\|_2^2\Big].
\end{aligned}
\end{equation}
This surrogate admits a natural interpretation as a weighted diffusion ELBO~\cite{elbo, vdm}, making it both principled and tractable for optimization. Consequently, the importance ratio becomes 
\begin{equation}
    \begin{aligned}
        \rho_i^{\boldsymbol{\theta}} = \frac{\pi^{\boldsymbol{\theta}}(\mathbf{o}_i | c)}{\pi^{\boldsymbol{\theta}_{\text{old}}}(\mathbf{o}_i | c)} = & \exp(\log \pi^{\boldsymbol{\theta}}(\mathbf{o}_i \mid c) - \log \pi^{\boldsymbol{\theta}_{\text{old}}}(\mathbf{o}_i \mid c))\\ = & \exp(-\mathcal{L}_w(\boldsymbol{\theta}\mid \mathbf{o}_i, c) + \mathcal{L}_w(\boldsymbol{\theta}_{\text{old}}\mid \mathbf{o}_i, c)).
    \end{aligned}
\end{equation}
In practice, we approximate these ELBO-based surrogates using Monte Carlo estimation. Formally, we average the loss over $N_{\text{MC}}$ sampled timestep-noise pairs $\{(t_j, \boldsymbol{\epsilon}_j)\}_{j=1}^{N_{\text{MC}}}$:
\begin{equation}
    \hat{\mathcal{L}}_w(\boldsymbol{\theta}\mid \mathbf{o}_i, c) = \frac{1}{N_{\text{MC}}} \sum_{j=1}^{N_{\text{MC}}} \ell^{\boldsymbol{\theta}}_{i}(t_j, \boldsymbol{\epsilon}_j),
\end{equation}
where the individual loss term evaluated at each step is
\begin{equation}
    \ell^{\boldsymbol{\theta}}_{i}(t_j, \boldsymbol{\epsilon}_j) = w_{t_{j}}\left\| \texttt{NN}^{\boldsymbol{\theta}}_{t_{j}}(\mathbf{z}_{t_{j}}, c) - \mathbf{r}_{t_{j}}(\mathbf{o}_i, \boldsymbol{\epsilon}_j) \right\|_2^2.
    \label{eq:loss_term}
\end{equation}
We further detail the theoretical motivation for using these ELBO-based surrogates in~\cref{sec:motivation}.

Direct use of these surrogates leads to unstable training dynamics and suboptimal performance. In~\cref{sec:analysis}, we empirically analyze this failure and offer a plausible explanation that motivates three effective \textcolor{Plum}{surrogate variance reduction} techniques, detailed in~\cref{sec:techniques1}. These techniques are used consistently throughout all experiments. Furthermore, in~\cref{sec:techniques2}, we distill best practices for three representative \textcolor{ForestGreen}{gradient step regulation} techniques, selecting and applying the most effective technique for each training scenario. The complete algorithm is outlined in Alg.~\ref{alg:vgrpo}, with the key techniques highlighted.

\subsection{Motivation}
\label{sec:motivation}
From a variational perspective, diffusion models~\cite{diffusion, ddpo} are pretrained by maximizing the ELBO on the model log-likelihoods of data samples~\cite{ddpm, diffusion}
\begin{equation}
    \text{ELBO}^{\boldsymbol{\theta}}(\mathbf{x}) \le \log\pi^{\boldsymbol{\theta}}(\mathbf{x}).
\end{equation}
For continuous-time diffusion models, the ELBO admits a simplified form~\cite{vdm, maximum, elbo}
\begin{equation}
\text{ELBO}^{\boldsymbol{\theta}}(\mathbf{x})
= -\frac{1}{2}\mathbb{E}_{t, \pi_1(\boldsymbol{\epsilon})}
\Big[-\frac{d\lambda_t}{dt}\Big\|\boldsymbol\epsilon^{\boldsymbol{\theta}}(\mathbf{z}_t) - \boldsymbol{\epsilon} \Big\|_2^2 \Big] + C,
\label{eq:elbo}
\end{equation}
where $\lambda_t = \log(\tfrac{a_t^2}{b_t^2})$ is the log signal-to-noise ratio (log-SNR), $\boldsymbol{\epsilon}^{\boldsymbol{\theta}}(\mathbf{z}_t)$ denotes the $\boldsymbol{\epsilon}$-prediction reparameterized from the model prediction $\texttt{NN}^{\boldsymbol{\theta}}_t(\mathbf{z}_t)$, and $C$ is constant w.r.t. model parameters $\boldsymbol{\theta}$~\cite{vdm, maximum}.

In practice, pretraining objectives augment \cref{eq:elbo} with a weighting function $w_t'$ to improve empirical performance:
\begin{equation}
\mathcal{L}_{w'}(\boldsymbol{\theta})
= \frac{1}{2}\mathbb{E}_{t, \pi_0(\mathbf{x}), \pi_1(\boldsymbol{\epsilon})}
\Big[w'_t \cdot -\frac{d\lambda_t}{dt}\Big\| \boldsymbol\epsilon^{\boldsymbol{\theta}}(\mathbf{z}_t) - \boldsymbol{\epsilon} \Big\|_2^2 \Big].
\end{equation}
This generalized form subsumes most common instantiations of~\cref{eq:loss}, making it compatible with a broad family of denoising generative models, such as rectified flow models~\cite{rf}. Although these models differ in their theoretical motivations and practical parameterizations, they are fundamentally equivalent: under any fixed forward process, their marginal densities evolve according to the same Fokker–Planck equation~\cite{principles, sde}.

This equivalence forms the foundation of our approach. It bridges all these models with the variational diffusion formulation, allowing us to reinterpret their pretraining objectives as negative weighted diffusion ELBOs on the model log-likelihood~\cite{elbo, fpo}. This reinterpretation yields tractable surrogates for policy gradient optimization.

\subsection{Diagnosing Instability}
\label{sec:analysis}

Despite its elegance, this ELBO-based formulation has historically underperformed in visual generation~\cite{fpo, ddpo}. Our preliminary experiments confirm this limitation: a naive implementation within GRPO suffers from training instability and poor convergence (see~\cref{fig:ablation-main}).

\begin{figure}[!htbp]
    \centering
    \includegraphics[width=0.9\linewidth]{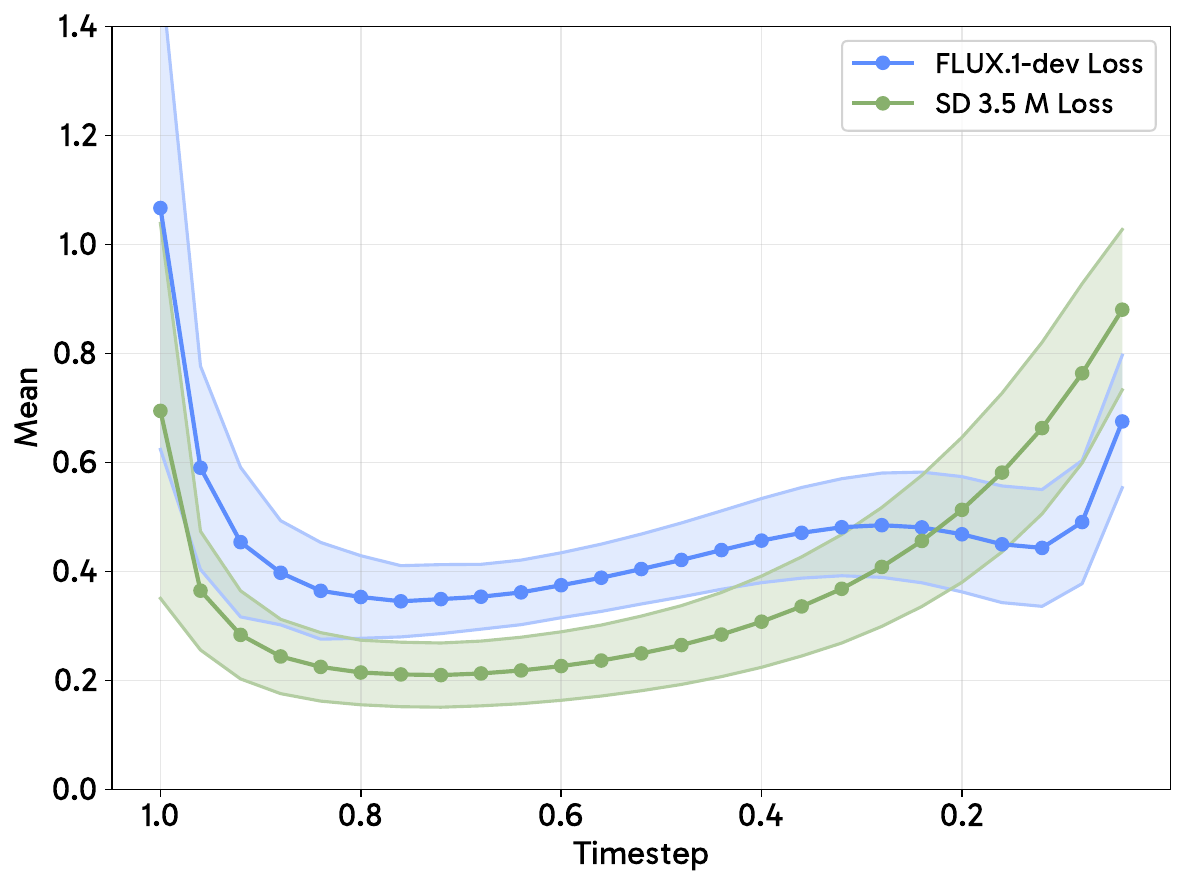}
    \vspace{-1em}
    \caption{\textbf{Per-sample loss varies substantially across timesteps. } Statistics are computed over 400 samples using~\cref{eq:loss_term}. Shaded regions indicate $\pm 1$ standard deviation.}
    \label{fig:analysis-1}
    \vspace{-1.5em}
\end{figure}

\begin{figure}[!t]
    \centering
    \includegraphics[width=\linewidth]{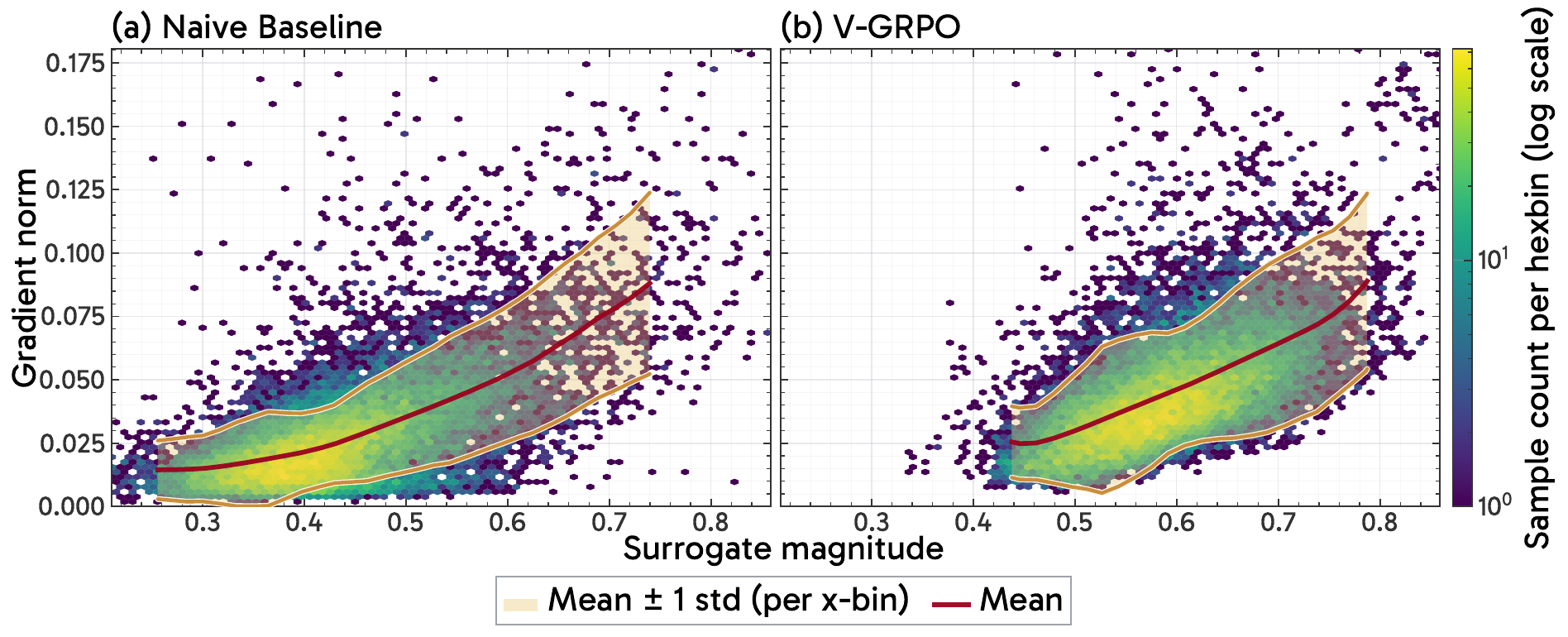}
    \vspace{-2em}
    \caption{\textbf{Gradient norms scale with surrogate magnitude.} Statistics are computed over $\sim$20K samples using FLUX.1-dev. Gradient norms are computed by backpropagating through the importance ratios, without applying clipping or scaling by the advantages. Mean curves are truncated to the 1st-99th percentile range.}
    \label{fig:analysis-2}
    \vspace{-1em}
\end{figure}

We hypothesize that this failure stems from excessive variance in the ELBO-based surrogates. As shown in~\cref{fig:analysis-1}, the magnitude of the per-sample loss $\ell^{\boldsymbol{\theta}}_{i}(t_j, \boldsymbol{\epsilon}_j)$ varies substantially across timesteps $t_j$. Sampling $\{(t_j, \boldsymbol{\epsilon}_j)\}_{j=1}^{N_{\text{MC}}}$ independently for each output yields high-variance surrogates that could fail to faithfully reflect the relative likelihoods. This problem is further compounded by the observation in~\cref{fig:analysis-2} that gradient norms scale with surrogate magnitude. Unstable pretraining losses thus propagate into high-variance gradients, where noise in the surrogate magnitudes overwhelms the reward signal and destabilizes training.

\subsection{Reducing Surrogate Variance}
\label{sec:techniques1}

Motivated by the preceding analysis, we identify three surrogate variance reduction techniques that prove effective.

\paragraph{Group-shared timestep-noise pairs.}

To reduce variance in the surrogate magnitude arising from random draws of $\{(t_j, \boldsymbol{\epsilon}_j)\}_{j=1}^{N_{\text{MC}}}$, we share these variables within each group. Concretely, for a given prompt $c$, we randomly sample a fixed set of timestep-noise pairs $\{(t_j, \boldsymbol{\epsilon}_j)\}_{j=1}^{N_{\text{MC}}}$ and apply this exact set across all outputs $\mathbf{o}_i$ generated from that prompt. By anchoring all outputs to the same stochastic basis, this design eliminates a major source of intra-group variance and renders the resulting policy gradient contributions directly comparable.

\paragraph{Stratified timestep sampling.}

Standard uniform timestep sampling can cause different outputs to draw from distinct regions of the noise schedule, leading to imbalanced optimization dynamics. To ensure representative and consistent coverage across the entire schedule for every output, we replace uniform sampling with a stratified scheme. Specifically, we partition the discretized timestep schedule into $N_{\text{MC}}$ disjoint, equal-length intervals and draw exactly one timestep $t_j$ from each interval when constructing the set of timestep-noise pairs. This guarantees uniform schedule coverage for each output and reduces surrogate variance attributable to timestep randomness.

\paragraph{Adaptive loss weighting.}

Following prior work~\cite{diffusionnft, dmd}, we employ an adaptive loss weighting scheme. First, we reparameterize the model output as an $\mathbf{x}$-prediction (e.g., $\mathbf{x}^{\boldsymbol{\theta}}(\mathbf{z}_t) = \mathbf{z}_t - t\texttt{NN}^{\boldsymbol{\theta}}_t(\mathbf{z}_t)$ for rectified flow) as this yields better performance (\cref{fig:pred}). We then apply a self-normalizing adaptive weighting function to this loss to yield the final objective
\begin{equation}
    \mathcal{L}^{\text{adaptive}}(\boldsymbol{\theta}\mid \mathbf{o}_i, c)
= \mathbb{E}_{t, \pi_1(\boldsymbol{\epsilon})}\left[ \frac{\left\Vert \mathbf{x}^{\boldsymbol{\theta}}(\mathbf{z}_t) - \mathbf{o}_i \right\Vert_2^2}{\operatorname{sg}(\frac{1}{d}\left\Vert\mathbf{x}^{\boldsymbol{\theta}}(\mathbf{z}_t) - \mathbf{o}_i\right\Vert_1)}\right],
\end{equation}
where $\operatorname{sg}(\cdot)$ denotes the stop-gradient operator and $d$ is the dimensionality of $\mathbf{o}_i$. Converting to $\mathbf{x}$-prediction implicitly places greater weight on higher noise levels, while self-normalization approximately aligns the gradient magnitudes across per-sample losses that vary in scale.

\paragraph{}Applying all techniques effectively reduces the variance in the gradient norm caused by the surrogate magnitude. First, the surrogate variance itself is reduced: the coefficient of variation (CV) of surrogate magnitude drops from 0.230 to 0.128, driven primarily by a lower mean within-group CV (0.170 $\rightarrow$ 0.038). Second, the gradient norm becomes less sensitive to surrogate magnitude, as reflected by a reduced coefficient of determination for a quadratic fit (0.406 $\rightarrow$ 0.328). These effects are evidenced in~\cref{fig:analysis-2}.

With this source of variance mitigated, these techniques significantly improve both stability and performance in actual training (see \cref{fig:ablation-main}).

\subsection{Controlling Gradient Steps}
\label{sec:techniques2}

Stable online RL requires careful control over gradient steps to prevent overly aggressive policy updates, a balance that is particularly crucial for our method. While numerous regularization and clipping strategies already exist in the literature, their effectiveness varies significantly depending on the training scenario. We distill best practices for three representative techniques, detailing how each can be optimally leveraged across different configurations in our framework.

\paragraph{Importance ratio clipping.}
Importance ratio clipping, introduced in PPO~\cite{ppo}, is a standard stabilization technique in policy gradient methods and, in practice, is \textit{generally sufficient to ensure stable training for our method in most settings}. Nonetheless, we find that two additional techniques could be beneficial in more specialized training scenarios.

\paragraph{KL penalty.} To penalize the policy for deviating excessively from a reference model $\pi^{\boldsymbol{\theta}_{\text{ref}}}$, a Kullback-Leibler (KL) divergence is typically incorporated into the training objective. To avoid the overhead of a separate reference model, we compute the KL penalty against the behavior policy $\pi^{\boldsymbol{\theta}_{\text{old}}}$, requiring us to store only $\ell^{\boldsymbol{\theta}_{\text{old}}}_i(t_j, \boldsymbol{\epsilon}_j)$ rather than another set of weights. For continuous diffusion models, this divergence also admits a simplified formulation
\begin{equation}
\begin{aligned}
    \mathbb{D}(\pi^{\boldsymbol{\theta}} \Vert \pi^{\boldsymbol{\theta}_{\text{old}}}) &= \mathbb{E}_{\pi^{\boldsymbol{\theta}}(\mathbf{o}_i|c)}\Big[\log\frac{\pi^{\boldsymbol{\theta}}(\mathbf{o}_i|c)}{\pi^{\boldsymbol{\theta}_{\text{old}}}(\mathbf{o}_i|c)}\Big] \\
    &= \frac{1}{2}\mathbb{E}_{\substack{t, \pi^{\boldsymbol{\theta}}(\mathbf{o}_i|c), \\ \pi_1(\boldsymbol{\epsilon})}}
\Big[\!-\!\frac{d\lambda_t}{dt}\Big\|\boldsymbol\epsilon^{\boldsymbol{\theta}}(\mathbf{z}_t) \!-\! \boldsymbol\epsilon^{\boldsymbol{\theta}_{\text{old}}}(\mathbf{z}_t) \Big\|_2^2 \Big] \!+\! C.
\end{aligned}
\end{equation}

Similar to the ELBO, it is common to apply a weighting function to this divergence~\cite{dmd, diffusiondpo}, yielding a flexible family of objectives. For consistency with prior techniques, we estimate this divergence using stored timestep-noise pairs in the following specific per-output form:
\begin{equation}
\mathbb{D}^{\text{simple}}_i(\pi^{\boldsymbol{\theta}}\Vert \pi^{\boldsymbol{\theta}_{\text{old}}}) = \mathbb{E}_{t, \pi_1(\boldsymbol{\epsilon})}
\Big[\Big\|\mathbf{x}^{\boldsymbol{\theta}}(\mathbf{z}_t) - \mathbf{x}^{\boldsymbol{\theta}_{\text{old}}}(\mathbf{z}_t)\Big\|_2^2 \Big].
\end{equation}
Although we compute this penalty in terms of the reparameterized $\mathbf{x}$-prediction for convenience, we find it equally effective in practice to compute it directly using $\mathbf{v}$-prediction.

As shown in~\cref{tab:kl}, KL penalty plays a key role in preserving capabilities acquired during earlier training. However, on its own it is not always sufficient to ensure stability. We observe that it cannot suppress loss spikes that destabilizes our FLUX.1-dev~\cite{flux} training (\cref{fig:irc-vs-kl}).

\begin{table*}[!htp]\vspace{-0.7em}
\centering
\footnotesize
\caption{\textbf{Evaluation results for multi-reward experiments on FLUX.1-dev.} All methods are trained for 300 iterations. Baseline results are sourced from their original papers~\cite{mixgrpo,branchgrpo}. $\text{NFE}_{\pi^{\boldsymbol{\theta}_{\text{old}}}}$ reports the sum of sampling steps and $N_{\text{MC}}$ (only for V-GRPO). \textbf{Bold}: best; \underline{Underline}: second best; * marks the frozen strategy proposed in MixGRPO~\cite{mixgrpo}.}
\begin{tabularx}{\linewidth}{lllCCCC}
\toprule
Method & 
$\text{NFE}_{\pi^{\boldsymbol{\theta}_{\text{old}}}}$& $\text{NFE}_{\pi^{\boldsymbol{\theta}}}$ &
HPS-v2.1 & 
PickScore & 
ImageReward & 
UnifiedReward \\
\midrule

FLUX.1-dev & — & —  & 0.313 & 0.227 & 1.088 & 3.370 \\
\midrule
\multirow{3}{*}{+ DanceGRPO}
& 25 & 4    & 0.334 & 0.225 & 1.335 & 3.374 \\
& 25 & 4$^*$ & 0.333 & 0.229 & 1.235 & 3.325 \\
& 25 & 14  & 0.356 & 0.233 & 1.436 & 3.397 \\

\midrule

+ BranchGRPO-WidPru         &13.68  &8.625  &0.364  &0.231     &1.609   &  3.383  \\
+ BranchGRPO-DepPru         &13.68  &8.625 &  \underline{0.369}  &0.235   &1.625    & 3.404  \\
+ BranchGRPO-Mix            &13.68  &4.25  &0.363  &0.230    &1.598   &   3.384\\
+ BranchGRPO                &13.68  &13.68  &0.363  &0.229  &1.603    &3.386   \\

\midrule
\multirow{2}{*}{+ MixGRPO-Flash}
& 8 & 4$^*$ & 0.357 & 0.232 & 1.624 & 3.402 \\
& 16 & 4 & 0.358 & 0.236 & 1.528 & 3.407 \\
\midrule

+ MixGRPO & 25 & 4 & 0.367 & 
\underline{0.237} & 1.629 & 3.418 \\

\midrule
\multirow{2}{*}{+ V-GRPO} & 16 + 4 & 4 & \textbf{0.372} & \textbf{0.241} & \underline{1.731} & \textbf{3.437} \\

& 25 + 4 & 4 & \textbf{0.372} & \textbf{0.241} & \textbf{1.749} & \underline{3.436} \\

\bottomrule
\end{tabularx}
\label{tab:main-flux}
\end{table*}

\begin{figure*}[!htbp]
    \centering
    \includegraphics[width=\linewidth]{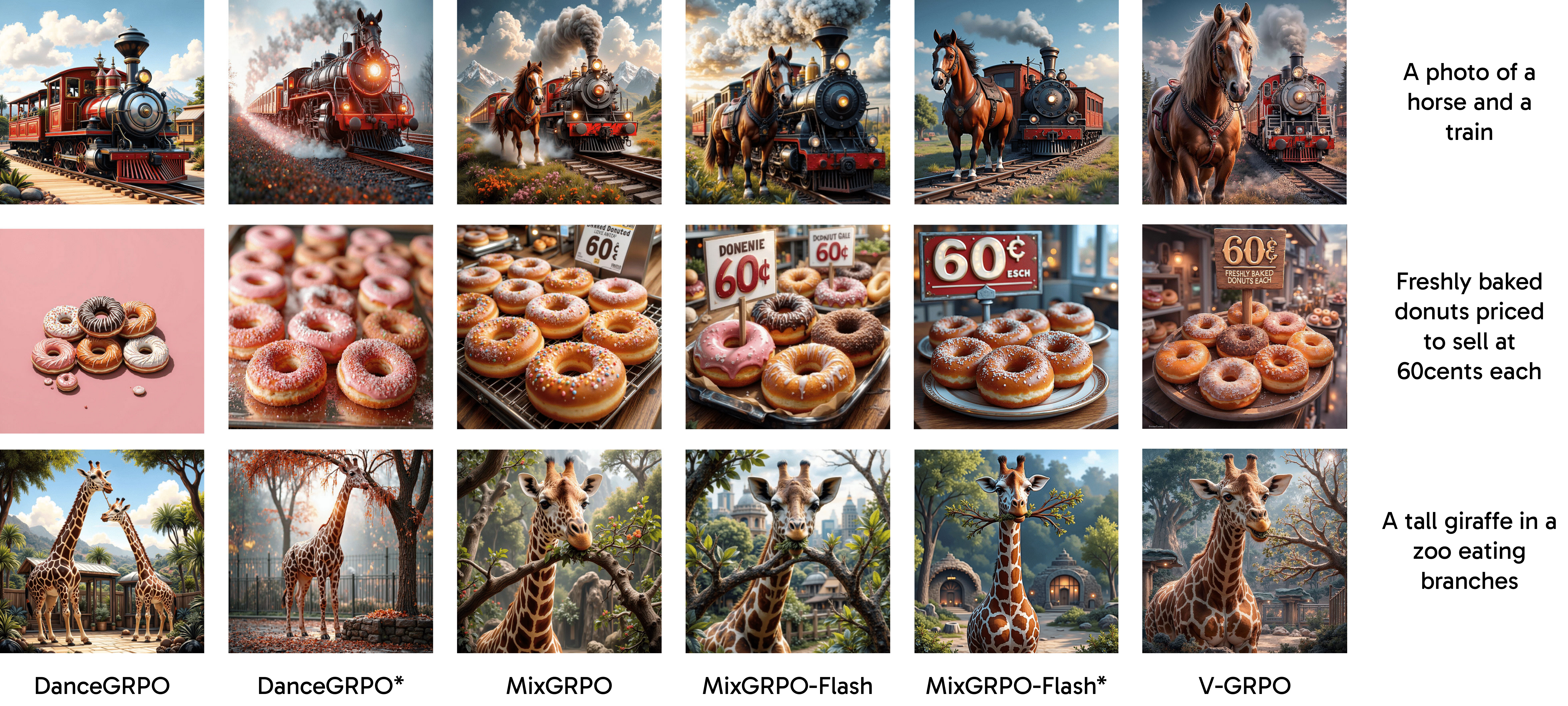}
    \vspace{-2em}
    \caption{\textbf{Qualitative comparison from the FLUX.1-dev main experiments.} V-GRPO achieves superior performance in alignment, coherence, and style. In the second example, it shows strong text rendering capabilities without leveraging task-specific rewards or datasets.}
    \label{fig:vis2}
    \vspace{-1.5em}
\end{figure*}

\paragraph{Advantage soft-clipping.}
When training in a fully on-policy manner (i.e., performing only 1 single gradient step per iteration), previously discussed techniques are no longer applicable (because $\boldsymbol{\theta} \equiv\boldsymbol{\theta}_{\text{old}}$). To address this, we propose to perform advantage soft-clipping based on the hyperbolic tangent, which remains applicable in this regime:
\begin{equation}
A^{\text{soft}} = \eta \cdot \tanh\left(\frac{1}{\eta}A\right),
\end{equation}
where $\eta$ is the clipping range. This preserves sensitivity for small advantages while smoothly bounding extreme values. 

As shown in~\cref{fig:adv-soft-clip}, this technique successfully stabilizes fully on-policy training. It also improves training stability under reduced sampling steps (see~\cref{fig:16-steps}). However,~\cref{fig:irc-vs-asc} illustrates that this approach can underperform prior methods in scenarios involving coarse-grained reward functions, such as GenEval~\cite{geneval}.

\paragraph{} The complete algorithm is presented in Alg.~\ref{alg:vgrpo}. In practice, we apply all surrogate variance reduction techniques while selectively enabling gradient step regularization based on the training configuration: \textit{KL penalty is used when preserving prior capabilities is critical; advantage soft-clipping is employed for fully on-policy training or when the number of sampling steps is limited; and importance ratio clipping is applied in most standard settings.} 

\section{Experiments}
\label{sec:experiments}

\subsection{Experiment Setup}
\label{sec:setup}

\colorlet{llgray}{lightgray!40}
\sethlcolor{llgray}

\begin{table*}[t]
    \centering
    \renewcommand{\arraystretch}{1.1}
    \vspace{-2mm}
    \footnotesize
    \caption{\textbf{Evaluation results for multi-stage, multi-reward experiments on SD 3.5 M.} Baseline results are sourced from DiffusionNFT~\cite{diffusionnft}. $\text{NFE}_{\pi^{\boldsymbol{\theta}_{\text{old}}}}$ reports the sum of sampling steps and $N_{\text{MC}}$ (for V-GRPO) or the ``number of timesteps optimized'' (for baselines). \hl{Gray-colored}: In-domain reward. $^\dagger$Evaluated under 1024$\times$1024 resolution. \textbf{Bold}: best; \underline{Underline}: second best.}
        \resizebox{\linewidth}{!}{\begin{tabular}{lccccccccccc}
            \toprule
            \multirow{2}{*}{Method}& \multirow{2}{*}{\#Steps} & \multirow{2}{*}{$\text{NFE}_{\pi^{\boldsymbol{\theta}_{\text{old}}}}$}& \multirow{2}{*}{$\text{NFE}_{\pi^{\boldsymbol{\theta}}}$} & \multicolumn{2}{c}{Rule-Based} & \multicolumn{6}{c}{Model-Based} \\ 
            \cmidrule(lr){5-6} \cmidrule(r){7-12} 
            && && GenEval & OCR  & PickScore & CLIPScore & HPSv2.1 & Aesthetics & ImgRwd & UniRwd \\ 
            \midrule
            SD XL$^\dagger$ & — & — & — & 0.55 & 0.14 & 0.2242 & 0.287 & 0.280 & 5.60 & 0.76 & 2.93 \\
            SD 3.5 L$^\dagger$ & — & — & — & 0.71 & 0.68 & 0.2291 & 0.289 & 0.288 & 5.50 & 0.96 & 3.25 \\
            \midrule
            SD 3.5 M (w/o CFG) & — & — & — & 0.24 & 0.12 & 0.2051 & 0.237 & 0.204 & 5.13 & -0.58 & 2.02 \\
            + CFG & — & — & — & 0.63 & 0.59 & 0.2234 & 0.285 & 0.279 & 5.36 & 0.85 & 3.03 \\
            \quad+ FlowGRPO & $>$5K & 40 & 40 & \cellcolor{llgray}\textbf{0.95} & 0.66 & 0.2251 & \underline{0.293} & 0.274 & 5.32 & 1.06 & 3.18 \\
             & 2K & 40 & 40 & 0.66 & \cellcolor{llgray}\textbf{0.92} & 0.2241 & 0.290 & 0.280 & 5.32 & 0.95 & 3.15 \\
             & 4K & 40 & 40 & 0.54 & 0.68 & \cellcolor{llgray}\underline{0.2350} & 0.280 & 0.316 & 5.90 & 1.29 & 3.37 \\
             + DiffusionNFT & 1.7K & 40 + 40 & 40 & \cellcolor{llgray}\underline{0.94} & \cellcolor{llgray}\underline{0.91} & \cellcolor{llgray}\textbf{0.2380} & \cellcolor{llgray}\underline{0.293} & \cellcolor{llgray}\underline{0.331} & \underline{6.01} & \underline{1.49} & \textbf{3.49} \\
             + V-GRPO & 580 & 40 + 6.9 & 6.9 & \cellcolor{llgray}0.91 & \cellcolor{llgray}\underline{0.91} & \cellcolor{llgray}\underline{0.2350} & \cellcolor{llgray}\textbf{0.298} & \cellcolor{llgray}\textbf{0.341} & \textbf{6.02} & \textbf{1.52} & \underline{3.43} \\
            \bottomrule
            \end{tabular}}
\vspace{-1.5em}
\label{tab:main-sd}
\end{table*}

\paragraph{Base models.}
We adopt two rectified flow models~\cite{rf} as our base models. Following DanceGRPO~\cite{dancegrpo}, MixGRPO~\cite{mixgrpo}, and BranchGRPO~\cite{branchgrpo}, we use FLUX.1-dev~\cite{flux}, a guidance-distilled model that operates without explicit Classifier-Free Guidance (CFG)~\cite{cfg}.

For comparison with Flow-GRPO~\cite{flowgrpo} and DiffusionNFT~\cite{diffusionnft}, we additionally employ Stable Diffusion 3.5 Medium (SD~3.5~M)~\cite{sd3}. Although this model typically relies on CFG for high-quality generation, we disable CFG in both training and evaluation. Consistent with prior findings~\cite{diffusionnft}, we observe that online RL effectively performs guidance distillation, eliminating the need for CFG.

\paragraph{Reward functions.} We employ two categories of reward functions: (1) rule-based rewards, including GenEval~\cite{geneval} for assessing compositional image–text alignment and optical character recognition (OCR)~\cite{paddleocr} for evaluating text rendering; and (2) model-based rewards, including HPSv2.1~\cite{hpsv2}, PickScore~\cite{pickscore}, CLIPScore~\cite{clipscore}, ImageReward~\cite{imagereward}, UnifiedReward~\cite{unifiedreward},  and Aesthetics~\cite{aesthetics}, which quantify image quality, image–text alignment, and alignment with human preferences.

\paragraph{Prompt datasets.} Following the baselines, all experiments on FLUX.1-dev are conducted using prompts from the HPDv2~\cite{hpsv2} dataset. For SD 3.5 M, we follow DiffusionNFT, using Flow-GRPO's prompt sets for GenEval and OCR experiments, and otherwise training on Pick-a-Pic~\cite{pickscore} with evaluation on DrawBench~\cite{drawbench}.

\paragraph{Training and evaluation.}

To ensure a fair comparison, our training and evaluation configurations follow those of the baselines. Unless otherwise stated, hyperparameter tuning is restricted to the number of gradient steps per iteration, importance ratio and advantage clipping ranges, KL coefficient, and number of timestep-noise pairs. Full implementation details are provided in~\cref{sec:impl}.

Our method decouples optimization from rollout, enabling the use of a second-order ODE sampler (DPMSolver++~\cite{dpmsolver++}) during rollout. While this precludes reusing model predictions from rollout for importance ratio computation and incurs a higher number of function evaluations (NFE) from $\pi^{\boldsymbol{\theta}_{\text{old}}}$, the overall framework remains competitively efficient. At evaluation time, we revert to a first-order Euler sampler to match baseline configurations.

For numerical stability, we employ BF16 mixed precision during rollout, while retaining full FP32 precision for master weights, ELBO-based surrogate computation for both $\pi^{\boldsymbol{\theta}_{\text{old}}}$ and $\pi^{\boldsymbol{\theta}}$, and the backward pass.

\subsection{FLUX.1-dev Main Results}

In our main experiments on FLUX.1-dev~\cite{flux}, we train the model for 300 iterations using an ensemble of four reward functions, including HPSv2.1~\cite{hpsv2}, PickScore~\cite{pickscore}, ImageReward~\cite{imagereward}, and UnifiedReward~\cite{unifiedreward}.

Quantitative comparisons against baselines are reported in~\cref{tab:main-flux}, while qualitative examples are illustrated in~\cref{fig:vis2} and~\cref{fig:vis1}. Our approach outperforms all baselines across every reward metric. Moreover, as shown in~\cref{tab:teaser}, V-GRPO converges $2\times$ faster than MixGRPO~\cite{mixgrpo}, reflecting substantially greater training efficiency.

\subsection{SD 3.5 M Main Results}

For SD 3.5 M experiments, we adopt the five-stage training curriculum from DiffusionNFT~\cite{diffusionnft}, running for 580 gradient updates with GenEval~\cite{geneval}, OCR~\cite{paddleocr}, HPSv2.1~\cite{hpsv2}, PickScore~\cite{pickscore}, and CLIPScore~\cite{clipscore}.

Quantitative comparisons against baselines are reported in~\cref{tab:main-sd}, while qualitative examples are illustrated in~\cref{fig:vis3}. Our approach matches the performance of DiffusionNFT~\cite{diffusionnft} while requiring roughly three times fewer gradient steps and a markedly lower NFE.

Moreover, V-GRPO achieves competitive performance in single-reward settings. Results are reported in~\cref{tab:single-sd}.

\begin{figure*}[htbp]
    \centering
    
    \begin{subfigure}[t!]{0.32\textwidth}
        \centering
        \vspace*{0pt}
        \includegraphics[width=\linewidth]{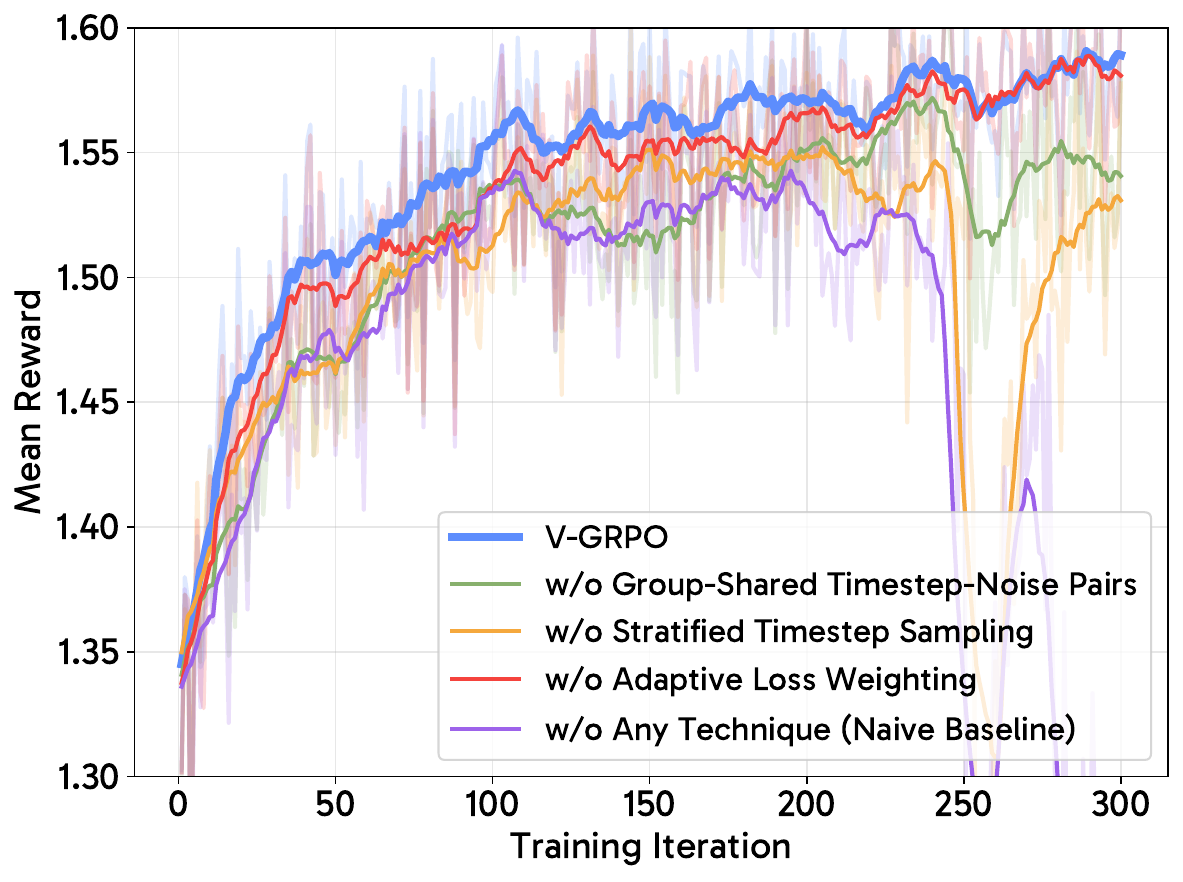}
        \caption{Reducing surrogate variance is essential for training FLUX.1-dev.}
        \label{fig:ablation-main}
    \end{subfigure}
    \hfill
    \begin{subfigure}[t!]{0.32\textwidth}
        \centering
        \vspace*{0pt}
        \includegraphics[width=\linewidth]{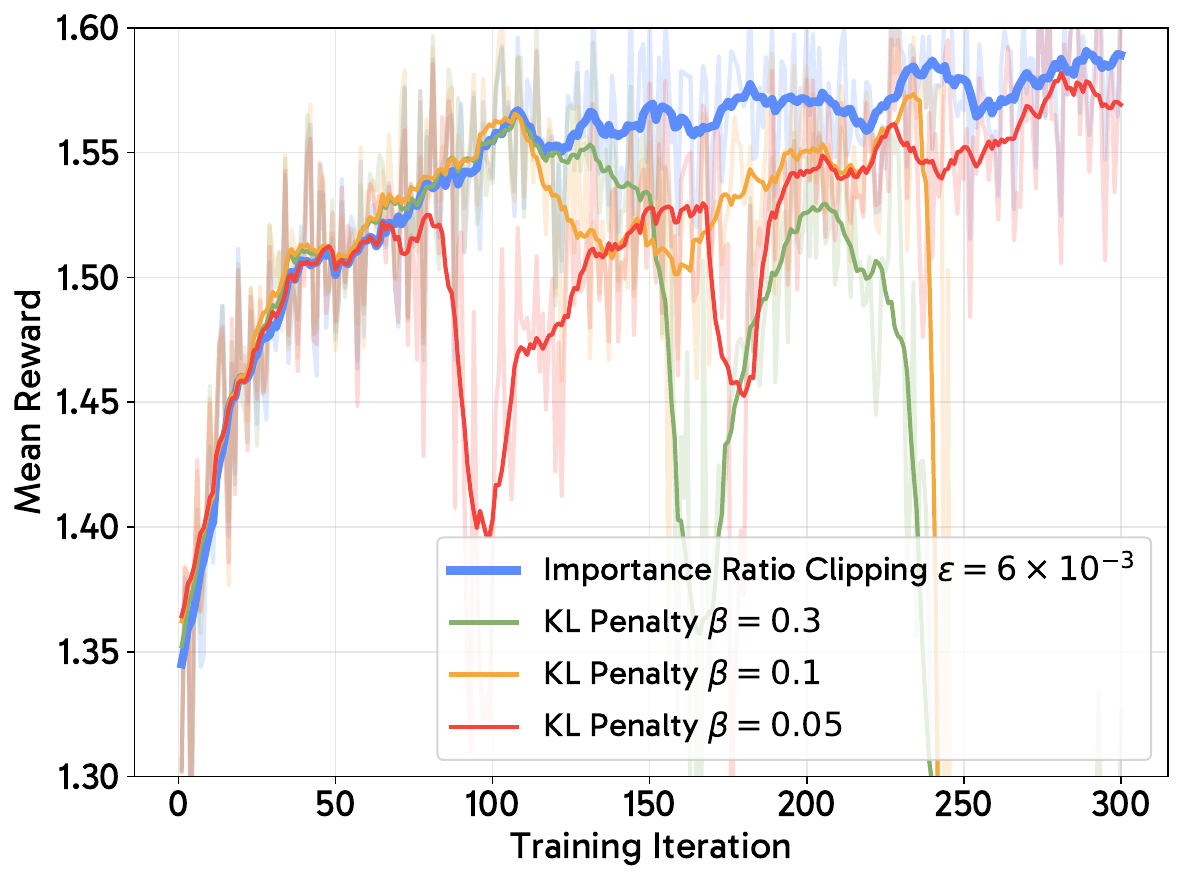}
        \caption{Importance ratio clipping, but not KL penalty, can resist loss spikes in FLUX.1-dev training.}
        \label{fig:irc-vs-kl}
    \end{subfigure}
    \hfill
    \begin{subfigure}[t!]{0.32\textwidth}
        \centering
        \vspace*{0pt}
        \includegraphics[width=\linewidth]{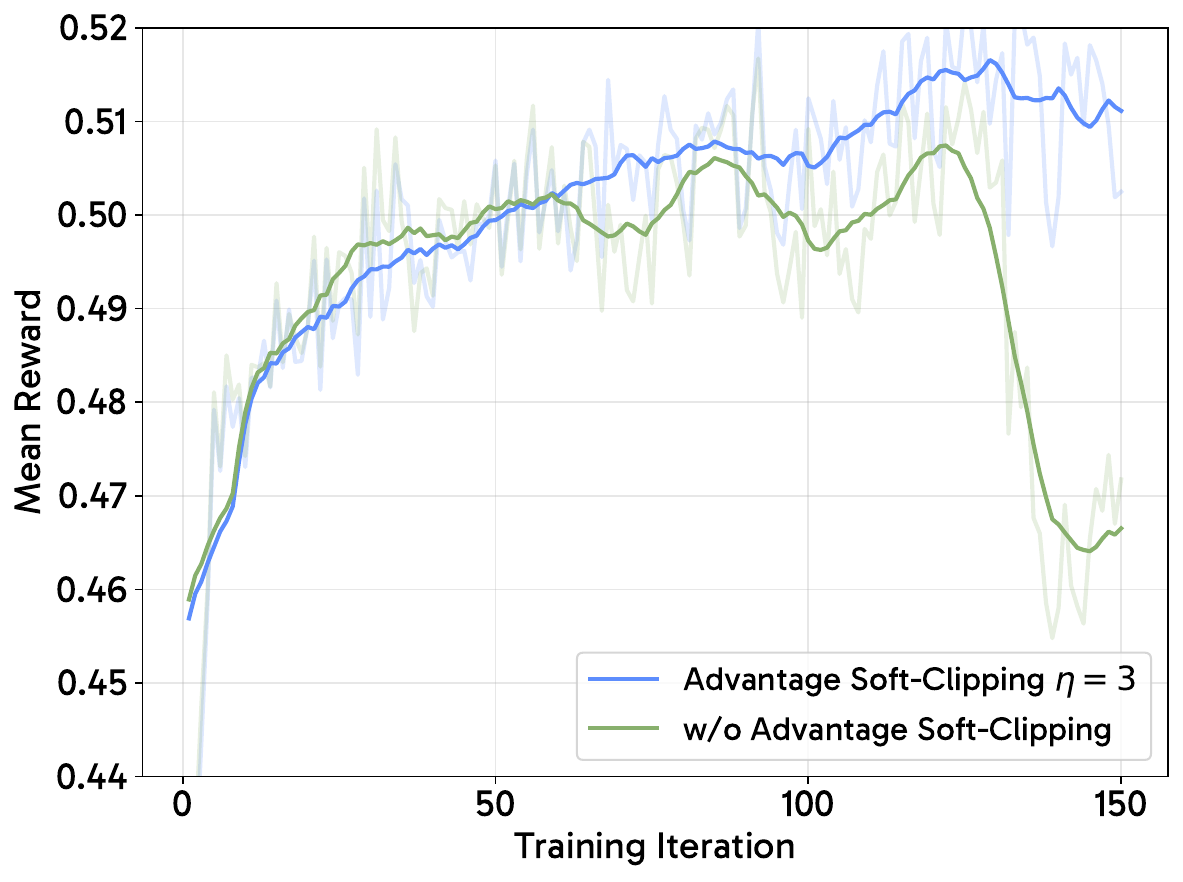}
        \caption{Advantage soft-clipping can stabilize the fully on-policy Stage-1 training of SD 3.5 M.}
        \label{fig:adv-soft-clip}
    \end{subfigure}

    \begin{subfigure}[t!]{0.32\textwidth}
        \centering
        \vspace*{0pt}
        \includegraphics[width=\linewidth]{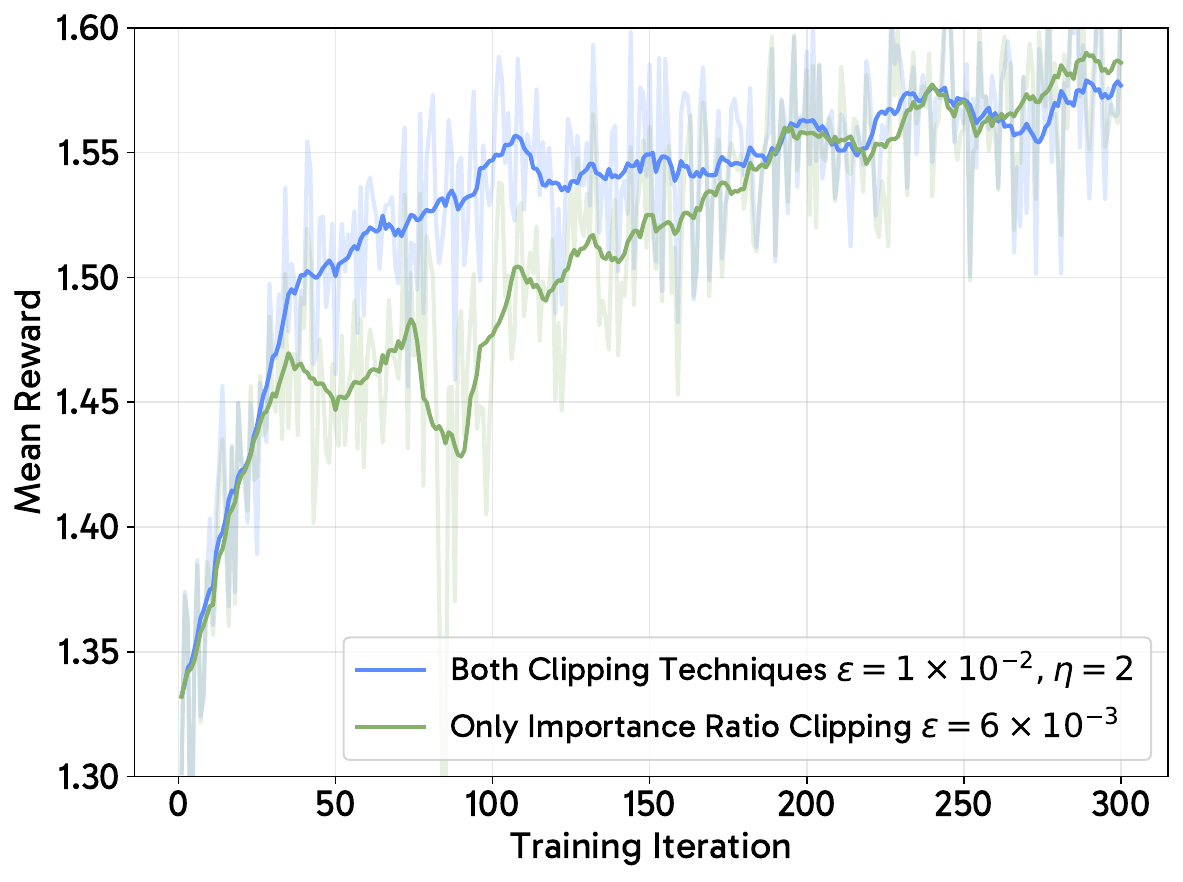}
        \caption{Advantage soft-clipping can further stabilize FLUX.1-dev training with 16 sampling steps.}
        \label{fig:16-steps}
    \end{subfigure}
    \hfill
    \begin{subfigure}[t!]{0.32\textwidth}
        \centering
        \vspace*{0pt}
        \includegraphics[width=\linewidth]{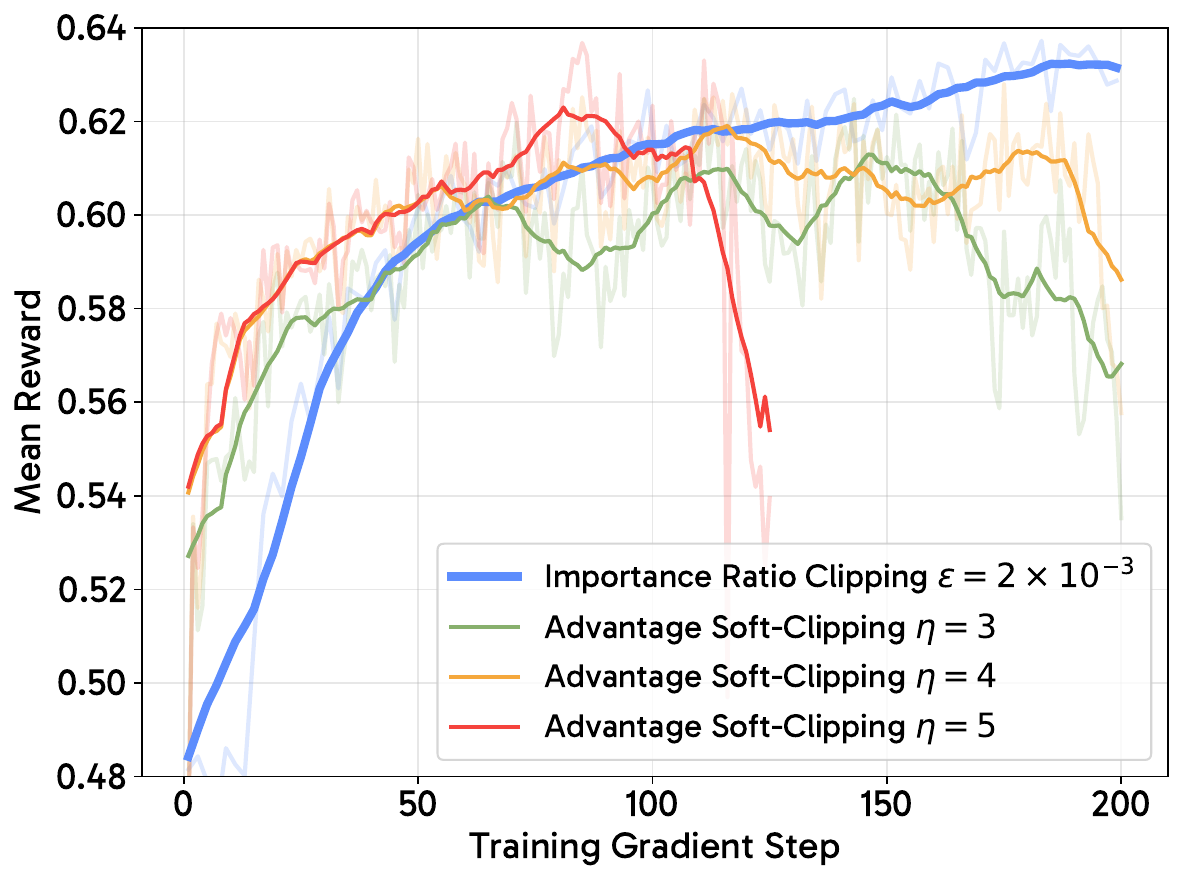}
        \caption{Advantage soft-clipping is suboptimal for Stage-2 training of SD 3.5 M that targets GenEval.}
        \label{fig:irc-vs-asc}
    \end{subfigure}
    \hfill
    \begin{subfigure}[t!]{0.32\textwidth}
        \centering
        \vspace*{0pt}
        \includegraphics[width=\linewidth]{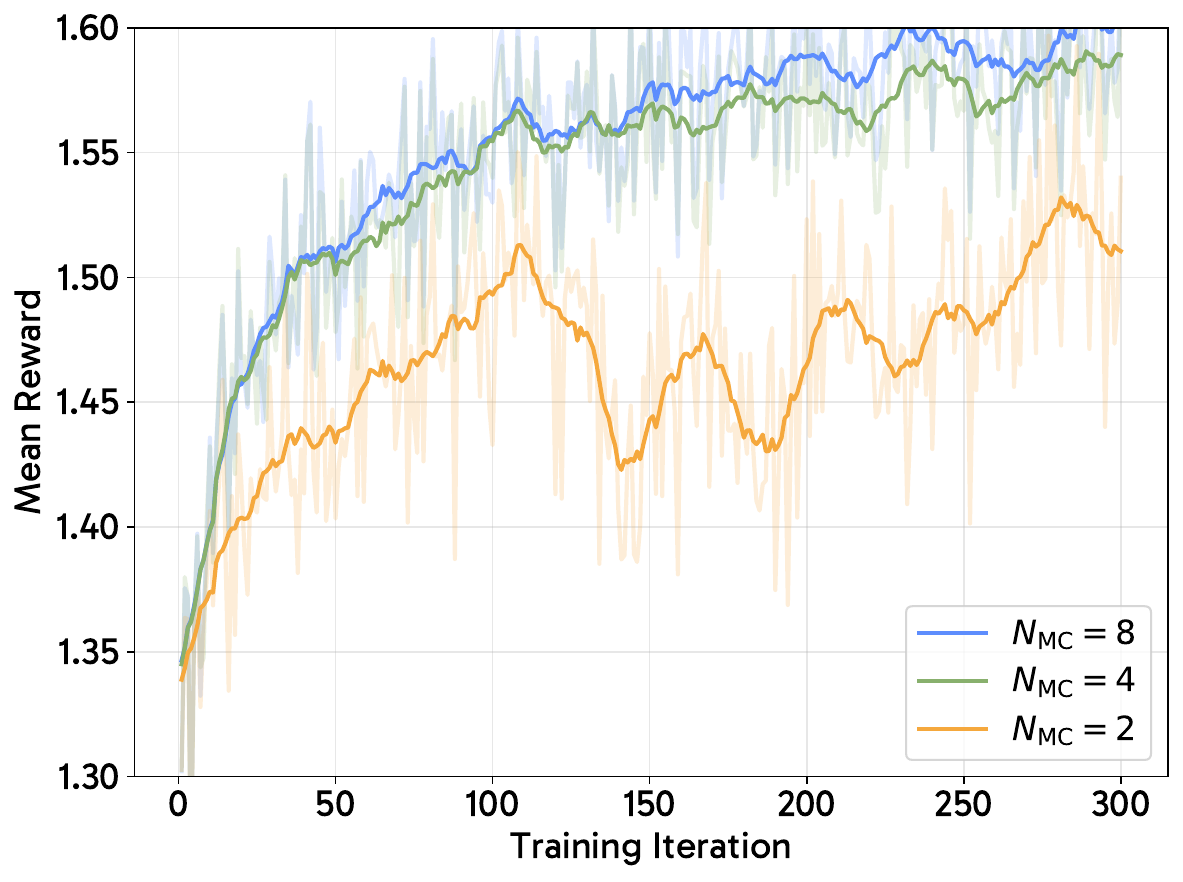}
        \caption{$N_{\text{MC}}$ saturates: too few samples hurt convergence, too many offer marginal gain.}
        \label{fig:ablation-nmc}
    \end{subfigure}

    \vspace{-0.5em}

    \caption{\textbf{Results for ablation studies.}}

    \vspace{-1.5em}
\end{figure*}

\subsection{Ablation Studies}

\paragraph{Reducing surrogate variance.} \cref{fig:ablation-main} ablates the proposed surrogate variance reduction techniques on FLUX.1-dev~\cite{flux}. Without these methods, the naive baseline suffers from severe training instability. Removing either group-shared timestep-noise pairs or stratified timestep sampling similarly destabilizes training, while omitting adaptive loss weighting causes a slight drop in performance. Together, all techniques are essential to achieve optimal results. 

\cref{fig:ablation-sd} shows that, while the naive baseline again suffers from severe instability during Stage-1 training of SD 3.5 M~\cite{sd3}, ablating individual components has a limited effect on both stability and performance. This indicates that SD 3.5 M is inherently more robust to ELBO-based training, so while our techniques collectively remain highly beneficial, no individual component is critical on its own in this case.

\begin{table}[htbp!]
    \centering
    \renewcommand{\arraystretch}{1.1}
    \vspace{-2mm}
    \footnotesize
    \caption{KL penalty preserves prior capabilities.}
    \vspace{-0.5em}
        \begin{tabular}{lcc}
            \toprule
            Method & GenEval& OCR\\
            \midrule
            Importance ratio clipping $\epsilon = 3\times10^{-2}$ & 0.87 & \textbf{0.93} \\
            KL penalty $\beta = 0.3$ & \textbf{0.91} & 0.91\\
            
            \bottomrule
            \end{tabular}
\vspace{-1.5em}
\label{tab:kl}
\end{table}

\paragraph{Controlling gradient steps.}
\cref{tab:kl} shows that a KL penalty effectively preserves prior capabilities during Stage-5 training of SD 3.5 M, which targets OCR~\cite{paddleocr} and does not use GenEval~\cite{geneval}, whereas importance ratio clipping induces a degradation in GenEval performance (0.92 $\rightarrow$ 0.87). However, \cref{fig:irc-vs-kl} indicates that the KL penalty alone is insufficient to stabilize training of FLUX.1-dev. In addition, while advantage soft-clipping improves stability of the fully on-policy Stage-1 training of SD 3.5 M (\cref{fig:adv-soft-clip}), and FLUX.1-dev training with sampling steps reduced from 25 to 16 (\cref{fig:16-steps}), \cref{fig:irc-vs-asc} shows that it underperforms importance ratio clipping with 2 gradient steps per iteration in Stage 2, which targets coarse-grained GenEval.

\paragraph{Effect of $N_{\text{MC}}$.} In \cref{fig:ablation-nmc}, we analyze the sensitivity to $N_{\text{MC}}$ on FLUX.1-dev. Reducing $N_{\text{MC}}$ from 4 to 2 prevents convergence, while increasing it to 8 yields only marginal gains. This mirrors findings in MDP-based methods regarding the ``number of timesteps optimized''~\cite{dancegrpo, mixgrpo}, suggesting a consistent saturation effect across both paradigms.
\section{Conclusion}
\label{sec:conclusion}

We present Variational GRPO (V-GRPO), an online RL method for denoising generative models that integrates ELBO-based surrogates into the GRPO algorithm, with simple techniques to reduce surrogate variance and control gradient steps. V-GRPO achieves state-of-the-art performance with substantial speedup over baselines. We hope these results help establish ELBO-based methods as the new default in this domain and inspire further research into their robustness and scalability.

\paragraph{Acknowledgments.} We thank Xinghan Li for insightful discussions and feedback on the manuscript. We also gratefully acknowledge Lambda, Inc. for providing partial computational support for this project. S.Y. is a Chan Zuckerberg Biohub — San Francisco Investigator.

{
    \small
    \bibliographystyle{ieeenat_fullname}
    \bibliography{main}

@String(CVPR= {IEEE Conf. Comput. Vis. Pattern Recog.})

@String(ICLR = {Int. Conf. Learn. Represent.})

@String(CVPR  = {CVPR})

@String(ICLR  = {ICLR})

@article{flowmatching,
  title={Flow matching for generative modeling},
  author={Lipman, Yaron and Chen, Ricky TQ and Ben-Hamu, Heli and Nickel, Maximilian and Le, Matt},
  journal={arXiv:2210.02747},
  year={2022}
}

@article{deepseekmath,
  title={Deepseekmath: Pushing the limits of mathematical reasoning in open language models},
  author={Shao, Zhihong and Wang, Peiyi and Zhu, Qihao and Xu, Runxin and Song, Junxiao and Bi, Xiao and Zhang, Haowei and Zhang, Mingchuan and Li, YK and Wu, Yang and others},
  journal={arXiv:2402.03300},
  year={2024}
}

@article{deepseekr1,
  title={Deepseek-r1: Incentivizing reasoning capability in llms via reinforcement learning},
  author={Guo, Daya and Yang, Dejian and Zhang, Haowei and Song, Junxiao and Zhang, Ruoyu and Xu, Runxin and Zhu, Qihao and Ma, Shirong and Wang, Peiyi and Bi, Xiao and others},
  journal={arXiv:2501.12948},
  year={2025}
}

@article{dpo,
  title={Direct preference optimization: Your language model is secretly a reward model},
  author={Rafailov, Rafael and Sharma, Archit and Mitchell, Eric and Manning, Christopher D and Ermon, Stefano and Finn, Chelsea},
  journal={NeurIPS},
  year={2023}
}

@article{ppo,
  title={Proximal policy optimization algorithms},
  author={Schulman, John and Wolski, Filip and Dhariwal, Prafulla and Radford, Alec and Klimov, Oleg},
  journal={arXiv:1707.06347},
  year={2017}
}

@article{dancegrpo,
  title={DanceGRPO: Unleashing GRPO on Visual Generation},
  author={Xue, Zeyue and Wu, Jie and Gao, Yu and Kong, Fangyuan and Zhu, Lingting and Chen, Mengzhao and Liu, Zhiheng and Liu, Wei and Guo, Qiushan and Huang, Weilin and others},
  journal={arXiv:2505.07818},
  year={2025}
}

@article{flowgrpo,
  title={Flow-grpo: Training flow matching models via online rl},
  author={Liu, Jie and Liu, Gongye and Liang, Jiajun and Li, Yangguang and Liu, Jiaheng and Wang, Xintao and Wan, Pengfei and Zhang, Di and Ouyang, Wanli},
  journal={arXiv:2505.05470},
  year={2025}
}

@article{mixgrpo,
  title={MixGRPO: Unlocking Flow-based GRPO Efficiency with Mixed ODE-SDE},
  author={Li, Junzhe and Cui, Yutao and Huang, Tao and Ma, Yinping and Fan, Chun and Yang, Miles and Zhong, Zhao},
  journal={arXiv:2507.21802},
  year={2025}
}

@article{dpmsolver++,
  title={Dpm-solver++: Fast solver for guided sampling of diffusion probabilistic models},
  author={Lu, Cheng and Zhou, Yuhao and Bao, Fan and Chen, Jianfei and Li, Chongxuan and Zhu, Jun},
  journal={Machine Intelligence Research},
  pages={1--22},
  year={2025},
  publisher={Springer}
}

@article{fpo,
  title={Flow Matching Policy Gradients},
  author={McAllister, David and Ge, Songwei and Yi, Brent and Kim, Chung Min and Weber, Ethan and Choi, Hongsuk and Feng, Haiwen and Kanazawa, Angjoo},
  journal={arXiv:2507.21053},
  year={2025}
}

@article{shortcut,
  title={Optimizing ddpm sampling with shortcut fine-tuning},
  author={Fan, Ying and Lee, Kangwook},
  journal={arXiv:2301.13362},
  year={2023}
}

@article{ddpo,
  title={Training diffusion models with reinforcement learning},
  author={Black, Kevin and Janner, Michael and Du, Yilun and Kostrikov, Ilya and Levine, Sergey},
  journal={arXiv:2305.13301},
  year={2023}
}

@article{dpok,
  title={Dpok: Reinforcement learning for fine-tuning text-to-image diffusion models},
  author={Fan, Ying and Watkins, Olivia and Du, Yuqing and Liu, Hao and Ryu, Moonkyung and Boutilier, Craig and Abbeel, Pieter and Ghavamzadeh, Mohammad and Lee, Kangwook and Lee, Kimin},
  journal={NeurIPS},
  year={2023}
}

@article{diffusiondpo,
  title={Diffusion model alignment using direct preference optimization},
  author={Wallace, Bram and Dang, Meihua and Rafailov, Rafael and Zhou, Linqi and Lou, Aaron and Purushwalkam, Senthil and Ermon, Stefano and Xiong, Caiming and Joty, Shafiq and Naik, Nikhil},
  journal={CVPR},
  year={2024}
}

@article{diffusion,
  title={Deep unsupervised learning using nonequilibrium thermodynamics},
  author={Sohl-Dickstein, Jascha and Weiss, Eric and Maheswaranathan, Niru and Ganguli, Surya},
  journal={ICML},
  year={2015},
}

@article{ddpm,
  title={Denoising diffusion probabilistic models},
  author={Ho, Jonathan and Jain, Ajay and Abbeel, Pieter},
  journal={NeurIPS},
  year={2020}
}

@article{rf,
  title={Flow straight and fast: Learning to generate and transfer data with rectified flow},
  author={Liu, Xingchao and Gong, Chengyue and Liu, Qiang},
  journal={arXiv:2209.03003},
  year={2022}
}

@article{cps,
  title={Coefficients-Preserving Sampling for Reinforcement Learning with Flow Matching},
  author={Wang, Feng and Yu, Zihao},
  journal={arXiv:2509.05952},
  year={2025}
}

@article{imagereward,
  title={Imagereward: Learning and evaluating human preferences for text-to-image generation},
  author={Xu, Jiazheng and Liu, Xiao and Wu, Yuchen and Tong, Yuxuan and Li, Qinkai and Ding, Ming and Tang, Jie and Dong, Yuxiao},
  journal={NeurIPS},
  year={2023}
}

@article{score,
  title={Score-based generative modeling through stochastic differential equations},
  author={Song, Yang and Sohl-Dickstein, Jascha and Kingma, Diederik P and Kumar, Abhishek and Ermon, Stefano and Poole, Ben},
  journal={arXiv:2011.13456},
  year={2020}
}

@article{diffusionnft,
  title={DiffusionNFT: Online Diffusion Reinforcement with Forward Process},
  author={Zheng, Kaiwen and Chen, Huayu and Ye, Haotian and Wang, Haoxiang and Zhang, Qinsheng and Jiang, Kai and Su, Hang and Ermon, Stefano and Zhu, Jun and Liu, Ming-Yu},
  journal={arXiv:2509.16117},
  year={2025}
}

@article{rwr,
  title={Aligning text-to-image models using human feedback},
  author={Lee, Kimin and Liu, Hao and Ryu, Moonkyung and Watkins, Olivia and Du, Yuqing and Boutilier, Craig and Abbeel, Pieter and Ghavamzadeh, Mohammad and Gu, Shixiang Shane},
  journal={arXiv:2302.12192},
  year={2023}
}

@article{cfg,
  title={Classifier-free diffusion guidance},
  author={Ho, Jonathan and Salimans, Tim},
  journal={arXiv:2207.12598},
  year={2022}
}

@article{salimans2022progressive,
  title={Progressive distillation for fast sampling of diffusion models},
  author={Salimans, Tim and Ho, Jonathan},
  journal={arXiv:2202.00512},
  year={2022}
}

@article{vdm,
  title={Variational diffusion models},
  author={Kingma, Diederik and Salimans, Tim and Poole, Ben and Ho, Jonathan},
  journal={NeurIPS},
  year={2021}
}

@article{maximum,
  title={Maximum likelihood training of score-based diffusion models},
  author={Song, Yang and Durkan, Conor and Murray, Iain and Ermon, Stefano},
  journal={NeurIPS},
  year={2021}
}

@article{elbo,
  title={Understanding diffusion objectives as the elbo with simple data augmentation},
  author={Kingma, Diederik and Gao, Ruiqi},
  journal={NeurIPS},
  year={2023}
}

@article{branchgrpo,
  title={Branchgrpo: Stable and efficient grpo with structured branching in diffusion models},
  author={Li, Yuming and Wang, Yikai and Zhu, Yuying and Zhao, Zhongyu and Lu, Ming and She, Qi and Zhang, Shanghang},
  journal={arXiv:2509.06040},
  year={2025}
}

@misc{flux,
    author={Black Forest Labs},
    title={FLUX},
    year={2024},
    howpublished={\url{https://github.com/black-forest-labs/flux}},
}

@article{hpsv2,
  title={Human preference score v2: A solid benchmark for evaluating human preferences of text-to-image synthesis},
  author={Wu, Xiaoshi and Hao, Yiming and Sun, Keqiang and Chen, Yixiong and Zhu, Feng and Zhao, Rui and Li, Hongsheng},
  journal={arXiv:2306.09341},
  year={2023}
}

@article{pickscore,
  title={Pick-a-pic: An open dataset of user preferences for text-to-image generation},
  author={Kirstain, Yuval and Polyak, Adam and Singer, Uriel and Matiana, Shahbuland and Penna, Joe and Levy, Omer},
  journal={NeurIPS},
  year={2023}
}

@article{unifiedreward,
  title={Unified reward model for multimodal understanding and generation},
  author={Wang, Yibin and Zang, Yuhang and Li, Hao and Jin, Cheng and Wang, Jiaqi},
  journal={arXiv:2503.05236},
  year={2025}
}

@article{sd3,
  title={Scaling rectified flow transformers for high-resolution image synthesis},
  author={Esser, Patrick and Kulal, Sumith and Blattmann, Andreas and Entezari, Rahim and M{\"u}ller, Jonas and Saini, Harry and Levi, Yam and Lorenz, Dominik and Sauer, Axel and Boesel, Frederic and others},
  journal={ICML},
  year={2024}
}

@article{adamw,
  title={Decoupled weight decay regularization},
  author={Loshchilov, Ilya and Hutter, Frank},
  journal={arXiv:1711.05101},
  year={2017}
}

@article{grpoguard,
  title={Grpo-guard: Mitigating implicit over-optimization in flow matching via regulated clipping},
  author={Wang, Jing and Liang, Jiajun and Liu, Jie and Liu, Henglin and Liu, Gongye and Zheng, Jun and Pang, Wanyuan and Ma, Ao and Xie, Zhenyu and Wang, Xintao and others},
  journal={arXiv:2510.22319},
  year={2025}
}

@article{awm,
  title={Advantage weighted matching: Aligning rl with pretraining in diffusion models},
  author={Xue, Shuchen and Ge, Chongjian and Zhang, Shilong and Li, Yichen and Ma, Zhi-Ming},
  journal={arXiv:2509.25050},
  year={2025}
}

@article{jit,
  title={Back to basics: Let denoising generative models denoise},
  author={Li, Tianhong and He, Kaiming},
  journal={arXiv preprint arXiv:2511.13720},
  year={2025}
}

@article{geneval,
  title={Geneval: An object-focused framework for evaluating text-to-image alignment},
  author={Ghosh, Dhruba and Hajishirzi, Hannaneh and Schmidt, Ludwig},
  journal={NeurIPS},
  year={2023}
}

@article{paddleocr,
  title={Paddleocr 3.0 technical report},
  author={Cui, Cheng and Sun, Ting and Lin, Manhui and Gao, Tingquan and Zhang, Yubo and Liu, Jiaxuan and Wang, Xueqing and Zhang, Zelun and Zhou, Changda and Liu, Hongen and others},
  journal={arXiv:2507.05595},
  year={2025}
}

@inproceedings{clipscore,
  title={Clipscore: A reference-free evaluation metric for image captioning},
  author={Hessel, Jack and Holtzman, Ari and Forbes, Maxwell and Le Bras, Ronan and Choi, Yejin},
  booktitle={EMNLP},
  year={2021}
}

@misc{aesthetics,
    author={Christoph Schuhmann},
    title={LAION-AESTHETICS},
    year={2022},
    howpublished={\url{https://laion.ai/blog/laion-aesthetics/}},
}

@article{lora,
  title={Lora: Low-rank adaptation of large language models.},
  author={Hu, Edward J and Shen, Yelong and Wallis, Phillip and Allen-Zhu, Zeyuan and Li, Yuanzhi and Wang, Shean and Wang, Liang and Chen, Weizhu and others},
  journal={ICLR},
  year={2022}
}

@article{drawbench,
  title={Photorealistic text-to-image diffusion models with deep language understanding},
  author={Saharia, Chitwan and Chan, William and Saxena, Saurabh and Li, Lala and Whang, Jay and Denton, Emily L and Ghasemipour, Kamyar and Gontijo Lopes, Raphael and Karagol Ayan, Burcu and Salimans, Tim and others},
  journal={NeurIPS},
  year={2022}
}

@incollection{sde,
  title={Stochastic differential equations},
  author={{\O}ksendal, Bernt},
  booktitle={Stochastic differential equations: an introduction with applications},
  year={2003},
  publisher={Springer}
}

@article{principles,
  title={The principles of diffusion models},
  author={Lai, Chieh-Hsin and Song, Yang and Kim, Dongjun and Mitsufuji, Yuki and Ermon, Stefano},
  journal={arXiv:2510.21890},
  year={2025}
}

@inproceedings{dmd,
  title={One-step diffusion with distribution matching distillation},
  author={Yin, Tianwei and Gharbi, Micha{\"e}l and Zhang, Richard and Shechtman, Eli and Durand, Fredo and Freeman, William T and Park, Taesung},
  booktitle={CVPR},
  year={2024}
}

@article{tempflowgrpo,
  title={Tempflow-grpo: When timing matters for grpo in flow models},
  author={He, Xiaoxuan and Fu, Siming and Zhao, Yuke and Li, Wanli and Yang, Jian and Yin, Dacheng and Rao, Fengyun and Zhang, Bo},
  journal={arXiv:2508.04324},
  year={2025}
}
}

\clearpage

\appendix
\section{Additional Implementation Details}
\label{sec:impl}

Our implementation adheres closely to the baseline methods, with deviations limited to the key techniques described in~\cref{sec:techniques1} and ~\cref{sec:techniques2}.

\subsection{FLUX.1-dev Experiments}
\paragraph{Main experiments.} Our setup follows MixGRPO~\cite{mixgrpo}, training and evaluating on prompts drawn from the HPDv2~\cite{hpsv2} dataset. For the reward functions, we utilize an ensemble of HPSv2.1~\cite{hpsv2}, PickScore~\cite{pickscore}, ImageReward~\cite{imagereward}, and UnifiedReward~\cite{unifiedreward}. PickScore is normalized in the same way as in MixGRPO~\cite{mixgrpo}. Multi-reward advantages are computed by averaging the mean of the individual reward advantages.

We optimize the model using AdamW~\cite{adamw} with a learning rate of $1 \times 10^{-5}$ and a weight decay of $1 \times 10^{-4}$. Training proceeds for 300 iterations, each comprising 4 gradient steps with a global batch size of 8 per step and a group size of 12. We do not maintain an exponential moving average (EMA) of weights during training.

We set the number of timestep-noise pairs to $N_{\text{MC}}=4$. For training with 25 sampling steps during rollout, importance ratios are clipped to $6 \times 10^{-3}$. For training with 16 sampling steps, importance ratios are clipped to $1 \times 10^{-2}$ and advantages are soft-clipped to 2. No KL penalty is applied in either configuration. 

During training rollout, we use a resolution of $720 \times 720$. Main experiments are conducted with both 16 and 25 sampling steps. At evaluation time, we scale this to 50 steps at a $1024 \times 1024$ resolution. To mitigate reward hacking while preserving strong generation quality during evaluation, we adopt the MixGRPO~\cite{mixgrpo} hybrid sampling strategy. Specifically, the trained model is used for the first $p_{\text{mix}}T$ steps (with $p_{\text{mix}} = 0.8$), and the original base model completes the remainder.

\paragraph{Ablation studies.} All ablation studies use 25 sampling steps. All other configurations are consistent with those used in the main experiments.

\subsection{SD 3.5 M Experiments}

\paragraph{Main experiments.} We adopt a multi-stage training curriculum from DiffusionNFT~\cite{diffusionnft}, which leverages diverse reward functions and prompt datasets. For multi-reward advantage estimation, we first aggregate individual rewards via averaging, and then compute advantages using these aggregated values.

We optimize the model using LoRA~\cite{lora} ($r = 32, \alpha = 64$) and the AdamW~\cite{adamw} optimizer with a learning rate of $3 \times 10^{-4}$ and a weight decay of $1 \times 10^{-4}$. Across all stages, training is conducted with a global batch size of 48 per gradient step and a group size of 24, matching the per-step configuration of DiffusionNFT~\cite{diffusionnft}.

 For consistency with DiffusionNFT, we prioritize using fully on-policy training with advantage soft-clipping, falling back to importance ratio clipping or a KL penalty when this is insufficient to achieve optimal performance. Unlike DiffusionNFT, we avoid reusing optimizer states between stages and do not maintain an EMA of weights.

Our multi-stage training curriculum is detailed below:
\begin{itemize}
 \item \textbf{Stages 1 and 3.} The model is trained on Pick-a-Pic~\cite{pickscore} using an ensemble of HPSv2.1~\cite{hpsv2}, PickScore~\cite{pickscore}, and CLIPScore~\cite{clipscore}. PickScore is normalized in the same way as in DiffusionNFT~\cite{diffusionnft}. In line with the baselines, each iteration performs only 1 gradient step. In both stages, we run 150 iterations (150 gradient steps), use $N_{\text{MC}} = 4$ timestep–noise pairs, and soft-clip advantages to 3. No importance ratio clipping or KL penalty is applied.

\item \textbf{Stages 2 and 4.} We add GenEval~\cite{geneval} to the three initial rewards. To improve performance through importance ratio clipping, each iteration performs 2 gradient steps. In Stage 2, training runs for 100 iterations (200 gradient steps) on Flow-GRPO~\cite{flowgrpo}'s prompt set with importance ratios clipped to $2 \times 10^{-3}$. In Stage 4, training runs for 25 iterations (50 gradient steps) with importance ratios clipped to $4 \times 10^{-3}$. In both stages, we use $N_{\text{MC}} = 10$. No KL penalty or advantage soft-clipping is applied.

\item \textbf{Stage 5.} We add OCR~\cite{paddleocr} to the three initial rewards. To preserve capabilities acquired during prior training via KL penalty, each iteration performs 2 gradient steps. Training runs for 15 iterations (30 gradient steps) on Flow-GRPO~\cite{flowgrpo}'s prompt set with $N_{\text{MC}} = 10$ and a KL coefficient of 0.3. No importance ratio clipping or advantage soft-clipping is applied.
\end{itemize}

During both training rollout and evaluation time, we use 40 sampling steps at a resolution of $512 \times 512$. Beyond the reward functions used during training, we further evaluate the trained model using out-of-domain metrics, including CLIPScore~\cite{clipscore}, UnifiedReward~\cite{unifiedreward}, and Aesthetics~\cite{aesthetics}.

To demonstrate the superior efficiency of our method, we compare its gradient step counts and NFE against those of DiffusionNFT in~\cref{tab:steps}.

\begin{table}[!htbp]
\small
\centering
\vspace{-0.5em}
\footnotesize
\caption{\textbf{Comparison of gradient step counts and NFEs across training stages.}
Our method delivers a $3\times$ speedup over DiffusionNFT in gradient steps, while also requiring fewer function evaluations (NFE) per step on average. DiffusionNFT reports per-stage step counts as approximate values due to early stopping, whereas our counts are exact.}

\begin{tabularx}{\linewidth}{lCCCCCC}
\toprule
\multirow{2}{*}{Stage} & \multicolumn{2}{c}{DiffusionNFT} & \multicolumn{2}{c}{V-GRPO} \\
\cmidrule(lr){2-3} \cmidrule(lr){4-5} & \#Steps & NFE & \#Steps & NFE \\
\midrule
1 (Human Preferences) & 800 & 120 & 150 & 48 \\
2 (GenEval) & 300 & 120 & 200 & 60 \\
3 (Human Preferences) & 200 & 120 & 150 & 48 \\
4 (GenEval) & 200 & 120 & 50  & 60 \\
5 (OCR) & 100 & 120 & 30  & 60 \\
\midrule
Total & 1700 & 120 & 580 & 53.8 \\
\bottomrule
\end{tabularx}
\label{tab:steps}
\vspace{-1em}
\end{table}

\paragraph{Single-reward experiments.} In our GenEval single-reward experiments, hyperparameters follow those in the Stage 4 of the main experiments, with training running for 300 iterations (600 gradient steps). For OCR, training runs for 25 iterations (25 gradient steps), using $N_{\text{MC}} = 10$ with advantages soft-clipped to 4. For PickScore, training runs for 300 iterations (300 gradient steps), using $N_{\text{MC}} = 4$ with advantages soft-clipped to 3. Both OCR and PickScore experiments perform 1 gradient step per iteration. All other configurations are consistent with those used in the main experiments.

\paragraph{Ablation studies.} Unless otherwise stated, all implementation details are the same as the main experiments.

\section{Additional Results}

\begin{figure*}
    \centering
    \includegraphics[width=0.75\linewidth]{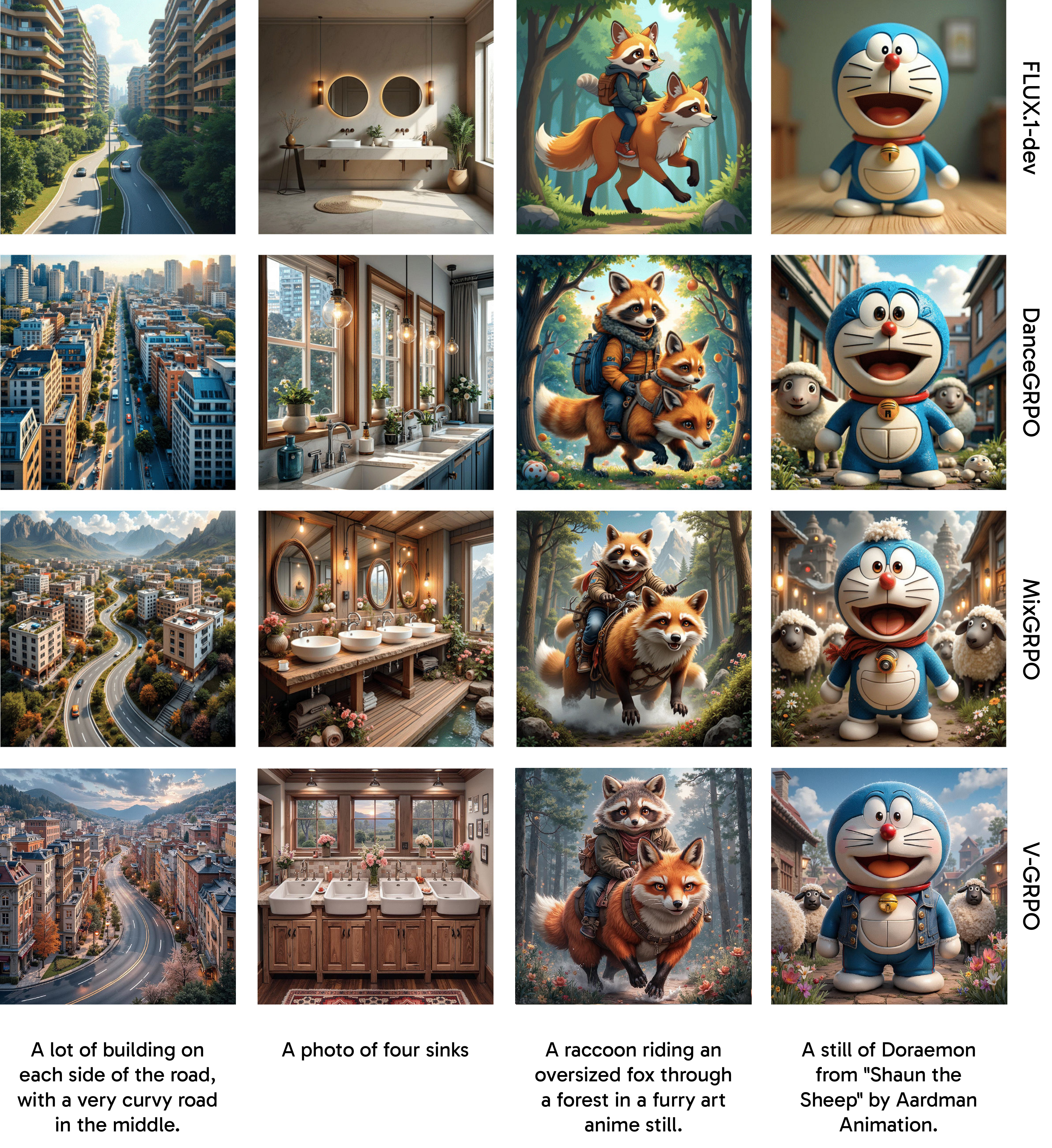}
    \caption{\textbf{Qualitative comparison from the FLUX.1-dev main experiments.} V-GRPO achieves superior performance in alignment, coherence, and style. In the fourth example, it demonstrates strong world knowledge.}
    \label{fig:vis1}
    \vspace{-0.5em}
\end{figure*}

\begin{figure*}[!htbp]
    \centering
    \includegraphics[width=\linewidth]{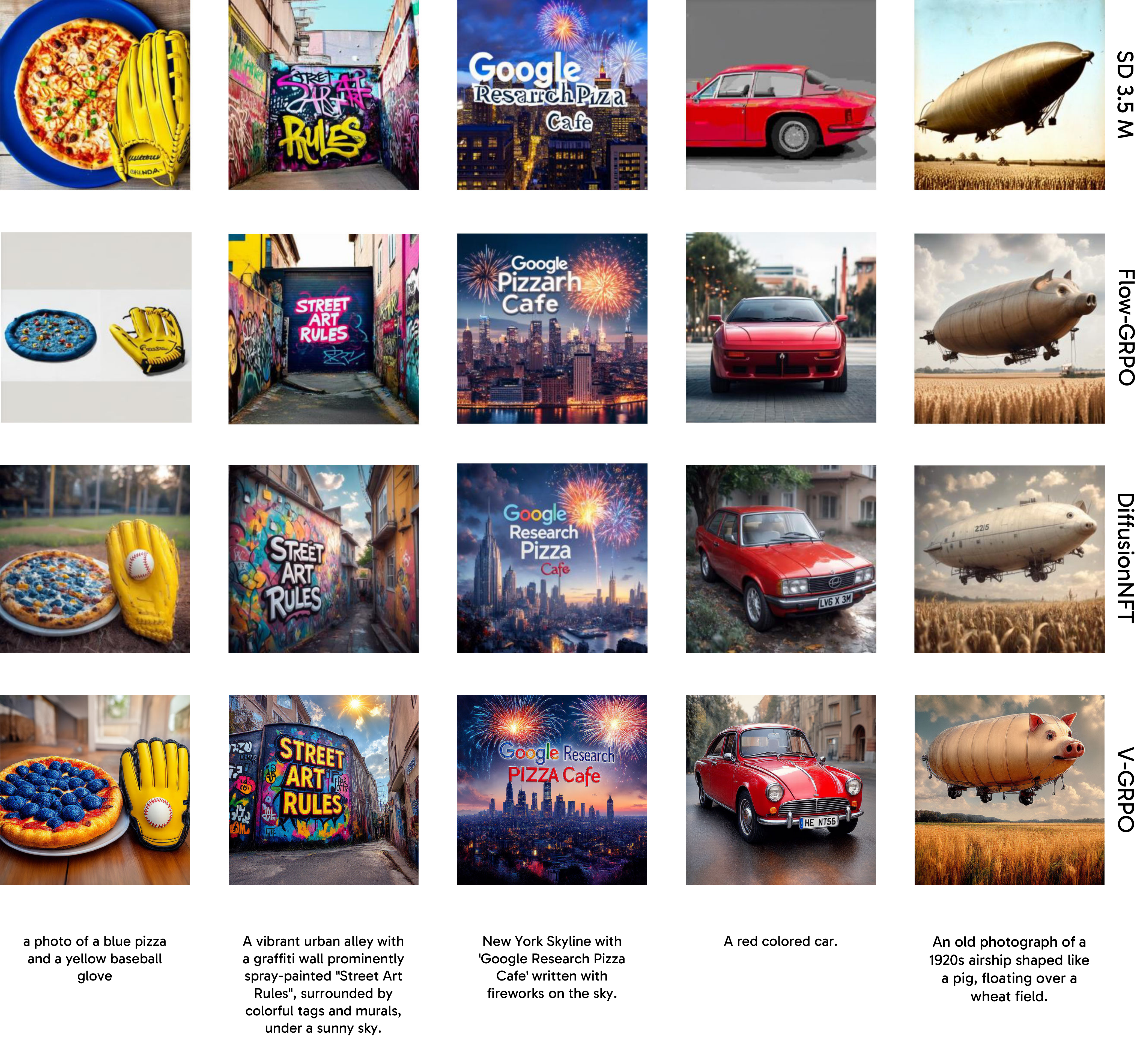}
    \vspace{-2em}
    \caption{\textbf{Qualitative comparison from the SD 3.5 M main experiments.} V-GRPO achieves superior performance in alignment, coherence, and style.}
    \label{fig:vis3}
    \vspace{-0.5em}
\end{figure*}

Additional qualitative examples from the FLUX.1-dev~\cite{flux} and SD 3.5 M~\cite{sd3} main experiments are illustrated in~\cref{fig:vis1} and~\cref{fig:vis3}, respectively.

Quantitative comparisons of single-reward experiments on SD 3.5 M are reported in~\cref{tab:single-sd}.

\begin{figure}[!htbp]
    \centering
    \includegraphics[width=\linewidth]{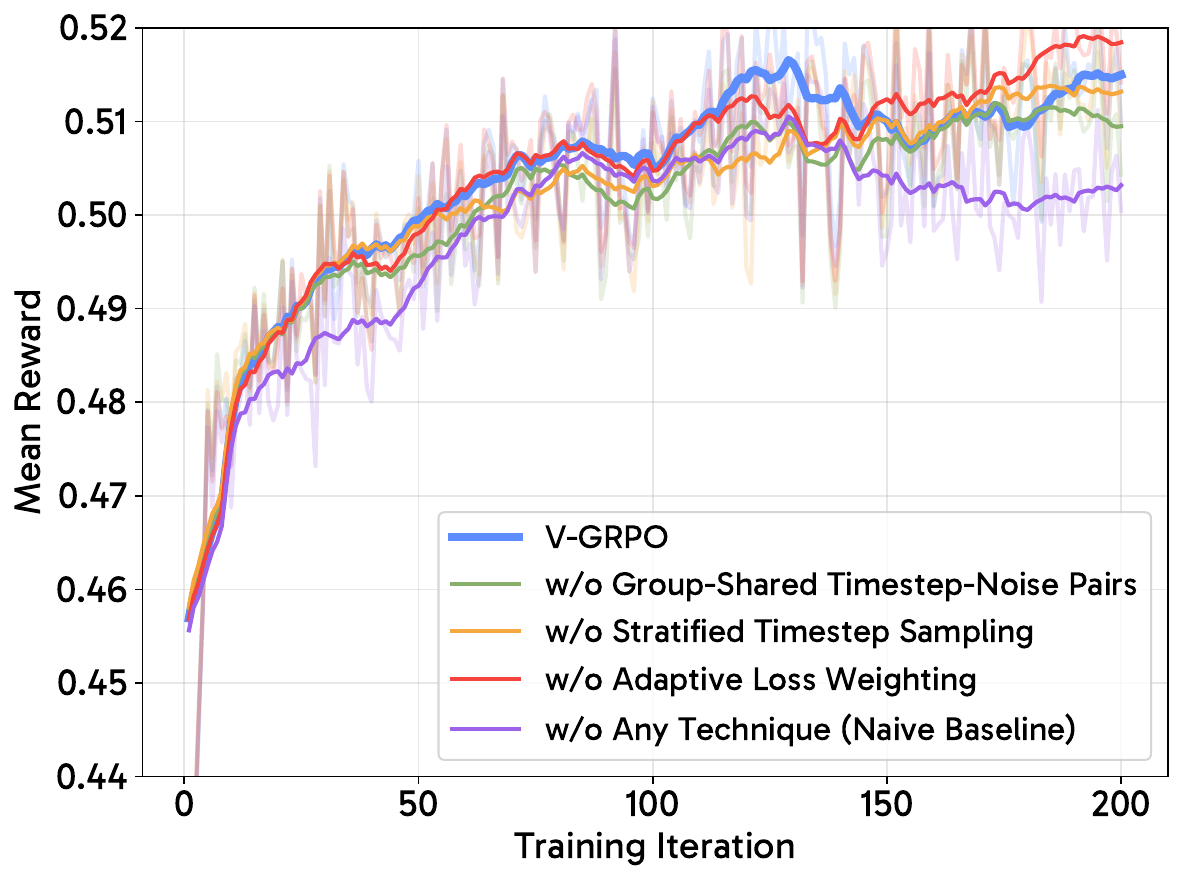}
    \vspace{-2em}
    \caption{\textbf{Ablation studies of surrogate variance reduction techniques.} Implementation details follow those of Stage-1 training in the SD 3.5 main experiments.}
    \label{fig:ablation-sd}
    \vspace{-0.5em}
\end{figure}

\begin{figure}[!htbp]
    \centering
    \includegraphics[width=\linewidth]{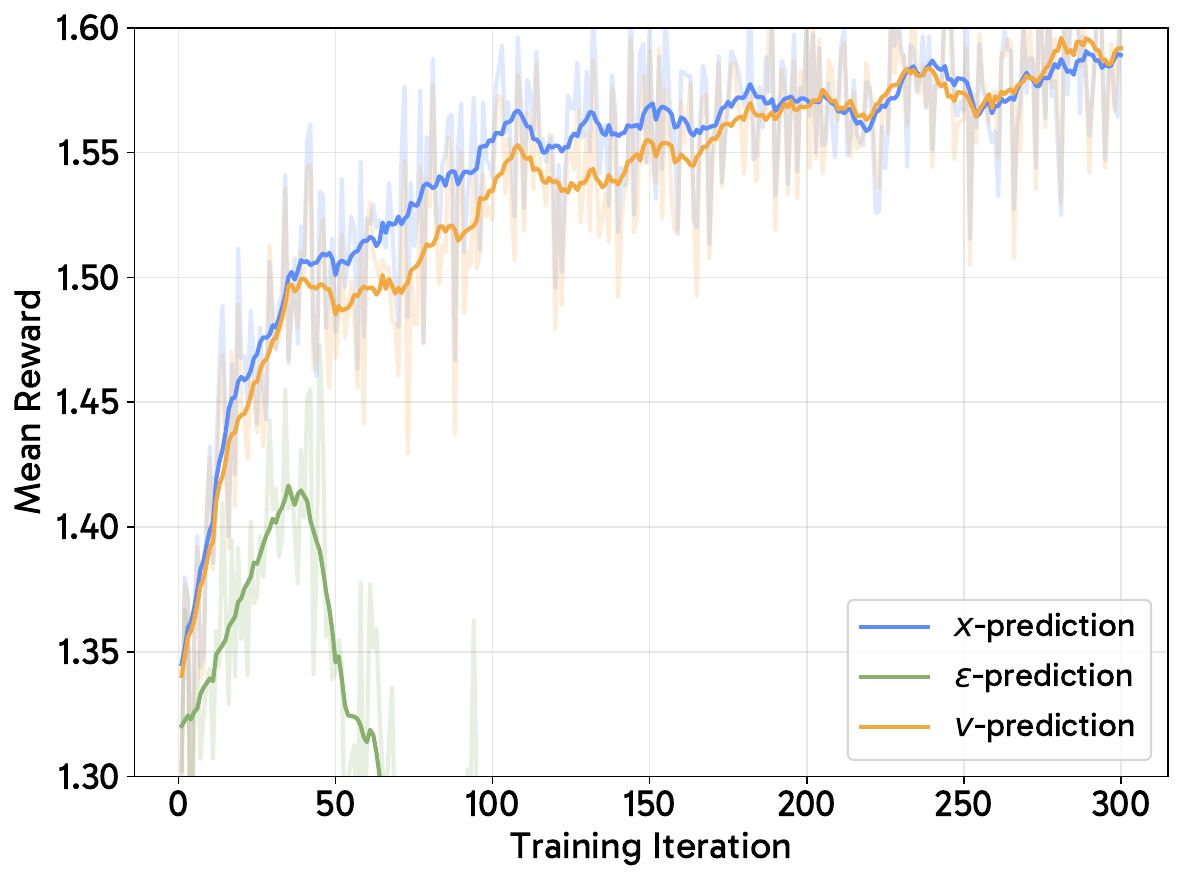}
    \vspace{-2em}
    \caption{\textbf{Ablation studies of alternative reparameterizations of model predictions.} Implementation details follow those used in the FLUX.1-dev experiments.}
    \label{fig:pred}
    \vspace{-1.5em}
\end{figure}

In~\cref{fig:ablation-sd}, we ablate the proposed surrogate variance reduction techniques on SD 3.5 M. While these techniques are collectively beneficial, no single component is individually critical. In~\cref{fig:pred}, we examine the effect of prediction parameterization on the adaptive loss weighting technique. $\boldsymbol{\epsilon}$-prediction leads to severe training collapse, whereas $\mathbf{v}$-prediction remains stable but yields slightly slower convergence than $\mathbf{x}$-prediction.

\begin{table*}[!htbp]
    \centering
    \renewcommand{\arraystretch}{1.1}
    \footnotesize
    \caption{\textbf{Evaluation results for single-reward experiments on SD 3.5 M.} All methods disable CFG during both training and evaluation. For models trained with the OCR reward, CFG is re-enabled when evaluating non-OCR rewards, following DiffusionNFT. Baseline results are sourced from DiffusionNFT. \hl{Gray-colored}: In-domain reward. \textbf{Bold}: best; \underline{Underline}: second best. } 
        \begin{tabular}{lccccccccc}
            \toprule
            \multirow{2}{*}{Method} & \multirow{2}{*}{\#Steps} & \multicolumn{2}{c}{Rule-Based} & \multicolumn{6}{c}{Model-Based} \\ 
            \cmidrule(lr){3-4} \cmidrule(r){5-10} 
            & & GenEval & OCR  & PickScore & CLIPScore & HPSv2.1 &Aesthetic & ImgRwd & UniRwd \\ 
            \midrule
            SD 3.5 M (w/o CFG) & — & 0.24 & 0.12 & 0.2051 & 0.237 & 0.204 & 5.13 & -0.58 & 2.02 \\
            + CFG & — & 0.63 & 0.59 & 0.2234 & 0.285 & 0.279 & 5.36 & 0.85 & 3.03 \\ \midrule
            + FlowGRPO & 4K & \cellcolor{llgray}\underline{0.97} & 0.30 & \underline{0.2178} & \underline{0.277} & \underline{0.248} & 5.15 & \textbf{0.74} & \underline{2.87} \\
            + DiffusionNFT & 1K & \cellcolor{llgray}\textbf{0.98} & \textbf{0.36} & \textbf{0.2192} & 0.271 & \textbf{0.251} & \textbf{5.33} & \underline{0.68} & \textbf{2.91} \\
             + V-GRPO & 600 & \cellcolor{llgray}\underline{0.97} & \underline{0.34} & 0.2150 & \textbf{0.280} & 0.225 & \underline{5.20} & 0.35 & 2.80 \\ \midrule
            + FlowGRPO & 1K & \textbf{0.66} & \cellcolor{llgray}0.96 & \textbf{0.2194} & \underline{0.280} & \textbf{0.257}& 5.18 & \underline{0.31} & \textbf{2.86} \\
            + DiffusionNFT & 150 & \underline{0.54} & \cellcolor{llgray}\underline{0.97} & 0.2163 & \textbf{0.281} & \underline{0.246} & \underline{5.19} & \textbf{0.37} & 2.81 \\ 
            + V-GRPO & 25 & 0.47 & \cellcolor{llgray}\textbf{0.98} & \underline{0.2170} & 0.277 & 0.243 & \textbf{5.21} & 0.28 & \underline{2.83} \\ \midrule
            + FlowGRPO & 4K & \underline{0.54} & 0.60 & \cellcolor{llgray}\underline{0.2362} & 0.257 & 0.295 & \textbf{6.42} & 1.17 & 3.17 \\
            + DiffusionNFT  & 2K & 0.53 & \textbf{0.64} & \cellcolor{llgray}\textbf{0.2403} & \textbf{0.270} & \textbf{0.315} & \underline{6.17} & \underline{1.29} & \textbf{3.40} \\
            + V-GRPO  & 300 & \textbf{0.66} & \underline{0.62} & \cellcolor{llgray}\textbf{0.2403} & \underline{0.267} & \underline{0.308} & \textbf{6.42} & \textbf{1.30} & \underline{3.26} \\
            \bottomrule
            \end{tabular}
\vspace{-2mm}
\label{tab:single-sd}
\end{table*}

\end{document}